\documentclass[review,1p]{elsarticle}

\usepackage{graphicx}
\usepackage{amsmath}
\usepackage{amssymb}
\usepackage{color}
\usepackage{algorithm} % algoritmi
\usepackage{algorithmic} % algoritmi
\usepackage{fixmath,fixltx2e,xspace}
\usepackage{bbm,booktabs}

\newcommand{\ra}[1]{\renewcommand{\arraystretch}{#1}}

\newcommand{\vone} 	{\mathbbm{1}}  

\newcommand{\ARE}{rotation synchronization\xspace}

\journal{International Journal of Computer Vision}
\begin{document}

\begin{frontmatter}

% udine
\author{Federica Arrigoni\corref{mycorrespondingauthor}}
\ead{arrigoni.federica@spes.uniud.it}
\cortext[mycorrespondingauthor]{Corresponding author}
\author{Andrea Fusiello}
\ead{andrea.fusiello@uniud.it}
\address{
Dipartimento di Ingegneria Elettrica, Gestionale e Meccanica,\\
University of Udine, Via Delle Scienze, 208 - 33100 Udine, Italy\\
}
%stm
\author{Beatrice Rossi}
\ead{beatrice.rossi@st.com}
\author{Pasqualina Fragneto}
\ead{pasqualina.fragneto@st.com}
\address{STMicroelectronics \\
Via Olivetti, 2  - 20864 Agrate Brianza, Italy\\}

\title{Robust Rotation Synchronization\\
 via Low-rank and Sparse Matrix Decomposition}

%%%%%%%%% ABSTRACT
\begin{abstract}
This paper deals with the \ARE
problem, which arises in global registration of 3D point-sets and in
structure from motion. The problem is formulated in an unprecedented way  as a  
``low-rank and 
sparse'' matrix decomposition that 
handles both outliers
and missing data. A minimization strategy, dubbed \textsc{R-GoDec}, 
is also proposed and evaluated experimentally against state-of-the-art 
algorithms on simulated and real data. The results show that \textsc{R-GoDec} is the fastest among the robust algorithms.

\end{abstract}

%%%%%%%%% BODY TEXT

\begin{keyword}
absolute rotations; global rotations; structure-from-motion;\ global
 registration;  $\ell_1$-regularization;  matrix completion; robust principal component analysis; low-rank $\&$ 
sparse matrix decomposition
\end{keyword}

\end{frontmatter}

%----------------------------------------------------------------------------------------

\section{Introduction}

In this paper we deal with the  \ARE problem, i.e the
problem of recovering the \emph{absolute} angular attitudes (rotations) -- with respect
to an external frame of reference -- of a set of local reference frames, given
their \emph{relative} angular attitudes. These local frames can be camera reference
frames, in which case we are in the context of structure from motion, or local
coordinates where 3D points are represented, in which case we are dealing with a
3D point-set registration problem.  

More abstractly, the goal of the \emph{group synchronization} problem \cite{Giridhar06,Singer2011} is to recover elements of a group 
from noisy measures of their ratios. 
In our case, absolute angular attitudes $R_1, \dots, R_n$ are elements of the Special Orthogonal Group $SO(3)$, and relative attitudes $R_{ij}=R_i R_j^T$ are their ratios.  The same problem is analysed in depth in
\cite{averaging}, under the name ``multiple rotation averaging''.

% Given a redundant
% number of \emph{relative} rotations $R_{ij}$ between frame pairs, the goal is to
% compute $N$ \emph{absolute} rotations $R_1, \dots, R_N$ that satisfy the compatibility 
% constraint $R_{ij}=R_i R_j^T$. 

%--------------Our contribution

Our solution to the \ARE problem is inspired by recent
advances in the fields of \emph{robust principal component analysis} and \emph{matrix completion}.  The main and original contribution 
of this paper is the formulation of the \ARE problem as a ``low-rank and sparse matrix decomposition'',  
by conceiving a  novel cost function that naturally includes missing data and 
outliers in its definition. Secondly, we develop a minimization scheme for that cost function -- called \textsc{R-GoDec} --
that leverages on the \textsc{GoDec} algorithm \cite{godec}. 

The resulting method is \emph{robust}, by construction,  \emph{fast}, thanks to the use of 
Bilateral Random Projections (BRP) in place of Singular Value Decomposition (SVD), and  \emph{compact}, as it consist of 
a single fixed point iteration that can be coded in a few lines of MATLAB. 
Most of all, the framework is modular, as --  in principle --  any low-rank and sparse decomposition method able to deal with 
outliers and missing data can replace \textsc{R-GoDec}.

%--------------Organizzazione del paper

This paper is organized as follows. 
Applications of the \ARE problem are presented in Section \ref{applications} while existing solutions are described in Section \ref{relatedwork}.
Section \ref{background} 
is an overview of the theoretical background required to define our
algorithm, i.e. low-rank and sparse matrix decomposition. 
Section \ref{formulation} defines the rotation synchronization problem.
Section \ref{our_method} provides a detailed description of our
robust solution. The method proposed in this
section is supported by experimental results on both synthetic and real data,
shown in Section \ref{experiments}. 
The conclusions %along with possible further developments 
are presented in Section \ref{conclusion}.
This paper is an extended version of \cite{RGodec}.

%----------------------------------------------------------------------------------------
\section{Applications}
\label{applications}

The synchronization problem over $SO(3)$ arises in 
different applications of Computer Vision, such as
structure from motion  and  multiple 3D point-set registration.
Other applications include sensor network localization and 
cryo-electron microscopy. 

%------------ 3D registration

\subsection{Multiple Point-set Registration}

The goal of multiple point-set registration  is to find the rigid transformations that bring multiple ($n \ge 2$) 3D point sets into alignment. 
Such point sets usually come from a 3D scanning device, which 
 can view only a fraction of an object from a given viewpoint.
Therefore, registration of multiple scans is  necessary to build a full 3D model of the object. 
Each rigid transformation is represented by an element of the Special Euclidean Group $SE(3)$, which is the semi-direct product of $SO(3)$ and $\mathbb{R}^3$.
In this paper we are interested only in the rotation component of the transformations.

Among the initial attempts to address this problem are the \emph{sequential} approaches introduced in \cite{Chen91,Pulli99}, that repeatedly register two point sets and integrate them into one model, until all the scans are considered. A well known failure case of these methods appears when the points sets are obtained using a turntable, since the constraint between the last and first scan is not used.

A different paradigm lies in \emph{global} methods, which are able to register simultaneously all the points sets. Such techniques take advantage of the redundancy in relative motions by using \emph{all} the constraints available between pairs of scans, thus they distribute the errors evenly across the scans, preventing drift in the solution. 
Global registration can be solved in
point (correspondences) space or in frame space.  In the first case, all the rotations are % WilBen01,BeiCro01
simultaneously optimized with respect to a cost function that depends on the
distance of corresponding points \cite{Pen96,KriLeeMoo07,BenShm98}. In the second case the optimization criterion is related to the
internal coherence of the network of rotations (and translations) applied to the
local coordinate frames \cite{FusCasRonMur02,Sharp02,Govindu14}.
% A related approach \cite{Thomas12} employs rank minimization for simultaneous alignment of range images, without computing matching points.

%------------ Structure from Motion

\subsection{Structure from Motion}

Recovering geometric information about a scene captured by multiple cameras has a great relevance in Computer Vision.
In the structure from motion problem such geometric information includes both scene structure, i.e. 3D coordinates of scene
points, and camera motion, i.e. absolute positions and attitudes of the cameras.
This problem appears also in the context of Photogrammetry under the name of  \emph{block orientation}.

Several systems have been developed to reconstruct large-scale scenes from a collection of unordered images.
They can be divided into three categories: \emph{structure-first}, \emph{structure-and-motion}, \emph{motion-first}.

Structure-first approaches (e.g. \cite{CroBei02}) begin with  estimating the structure and then recover the motion.
Specifically, stereo-models are built and co-registered, similarly to the point-set registration problem.

Structure-\emph{and}-motion techniques solve simultaneously for structure and motion.
Bundle block adjustment \cite{Triggs:bundle,Fusiello2015209}, resection-intersection methods \cite{Snavely06,BroLow05}, 
and hierarchical methods \cite{GheFarFus10,Ni12} belong to this category.
Although being highly accurate, these approaches suffer from two main disadvantages: on one hand they require intermediate expensive non-linear minimizations to contain error propagation, on the other hand the final reconstruction may depend on the order in which cameras are added or on the choice of the initial pair/triplet.

Motion-first methods initially recover the motion and then compute the structure \cite{Govindu01,MartinecP07,Arie12,moulon13,ozyesil2013stable}. 
They start from the relative motions determined from point matches among the images, they compute the angular attitude and position of the cameras with respect to an absolute coordinate frame, and then they return a sparse 
3D point cloud representing the scene. 
These motion-first methods are \emph{global}, for they take into account the entire relative information at
once, or, in other terms, they consider the whole \emph{epipolar graph}, where the nodes correspond to the cameras and the edges represent epipolar relationships. 
Global techniques have the advantage of fairly distributing the errors among the cameras, and thus they need bundle adjustment only at the end, thereby resulting in a reduction of the computational cost.
For this reason they have gained increasing attention in the community, similarly to frame-based methods for 3D point-set registration. 

%\todo{aggiugere il nostro dopo ICCV}

%With a fee exceptions \cite{Govindu10,eigse315}
With the exception of \cite{Govindu04}, most techniques split the motion estimation process in two stages. To begin with, the absolute rotation of each image is computed, then camera translations are recovered: we are concerned here with the first step only.

\subsection{Other applications}

The rotation synchronization problem arises also in the context of sensor network localization. In such a scenario the nodes of a sensor network can measure each other's relative orientations (by means of e.g. angle-of-arrival or pairwise distance sensing) with respect to their relative reference frames, and the goal is to express some other sensor measurements in a unique/global reference frame (measurements might include positions of targets, environment elements, etc.). Usually this application refers to planar networks, namely the synchronization problem in $SO(2)$ \cite{Cucuringu2012,piovan2013frame}.   
Another application regards structural biology. In \cite{Singer2011b} the problem of recovering the three-dimensional structure of a macromolecule from many cryo-electron microscopy (cryo-EM) images is considered. The direction from which each image is taken is unknown, and a rotation synchronization technique is used for determining the viewing direction of all cryo-EM images at once.

% SLAM?

%----------------------------------------------------------------------------------------

\section{Related work}
\label{relatedwork}

Several approaches have been proposed to solve the rotation synchronization problem, both in the context of multiple point-set registration and structure from motion (SfM). 
We shall divide them into non-robust and robust methods, according to the resilience  they show to rogue measures. In general, robustness is gained at the expense of statistical efficiency, i.e. non-robust estimators gets typically  closer to the Cram\`er-Rao bound \cite{Boumal2014}. On the other hand, non-robust methods can be skewed even by a single outlier, hence they are not applicable in practice unless they are preceded by an outlier detection stage.  

Outliers are very frequent when dealing with real data. In the SfM context,
for example, repetitive structures in the images cause mismatches which skew the
epipolar geometry. In global registration of 3D point-sets, outliers are caused
by faulty pairwise registration, which in turn may be originated  by insufficient
overlap and/or poor initialization.

% The former can handle a low-level of noise among the input rotations, but they are highly non robust and they can give wrong results in the presence of even a single outlier. For this reason, they need a preliminary step to detect and remove such outliers before performing rotation .
% The latter are inherently resilient to outliers since they incorporate robustness directly in the cost function, hence they are more efficient.
% A detailed survey on existing algorithms as well as a theoretical analysis of  rotation synchronization is reported in \cite{averaging}.

\subsection{Non-robust methods}

The authors of \cite{Sharp02} decompose the graph of neighbouring views into a set of cycles, and they propose an iterative procedure to recover the absolute 
rotations in which the error is distributed over these cycles.
As observed in \cite{Govindu14}, this technique performs a suboptimal set of averages and as a result it may converge to a local minimum.

In \cite{FusCasRonMur02,Govindu01} the rotation synchronization problem is cast to the optimization of an objective
function where rotations are parametrized as quaternions.  
Govindu in \cite{Govindu01} expresses the compatibility constraint between relative and absolute rotations as a linear system of equations which is solved in the least-squares sense,
while in \cite{FusCasRonMur02} the absolute rotations are computed using a quasi-Newton method.

The methods described in \cite{MartinecP07,Arie12,isprsarchives-XL-5-63-2014} perform $\ell_2$ averaging of relative rotations by using the chordal (Frobenius) distance. Without enforcing the orthogonality constraints, approximate solutions are computed, and they are subsequently projected onto $SO(3)$ by finding the nearest rotation matrices (in the Frobenius norm sense).
Martinec et al. in \cite{MartinecP07} compute a least-squares solution through vectorization and Singular Value Decomposition (SVD). This approach is extended in \cite{Arie12} using spectral decomposition or semi-definite programming.
A gradient descent method based on matrix completion is presented in \cite{isprsarchives-XL-5-63-2014}.
According to the analysis in \cite{MartinecP07}, methods involving matrices usually perform better than quaternion minimization.

A different approach consists in performing Lie-algebraic averaging in the group of 3D rotations \cite{Govindu04}.
This method exploits the Lie-group structure of $SO(3)$ and proposes an iterative scheme in which at each step the absolute rotations are updated by averaging relative rotations in the tangent space.

%The method in \cite{SinhaSS10} assumes as input vanishing points in addition to relative rotation estimates.

\subsection{Robust methods}

% quelli con outlier detection

The main drawback of the previous techniques is that they suffer from the presence of inconsistent relative rotations, and thus they need a preliminary step to detect and remove such outliers \emph{before} computing the absolute rotations.  

A wide overview of methods aimed at detecting outlier rotations can be found in \cite{moulon13}.
These approaches \cite{EnqvistKO11,OlssonE11a,Govindu06,bayesian} check for cycle consistency, i.e. deviation from identity, within the epipolar graph.
Enqvist et al. \cite{EnqvistKO11} consider a maximum-weight spanning tree, where the weight of an edge is the number of inlier correspondences, and they analyse cycles formed by the remaining edges. 
In \cite{isprsarchives-XL-5-63-2014} some heuristics based on cycle basis  are introduced to improve this scheme.
In \cite{bayesian} a Bayesian framework is used to classify all the edges into inliers and outliers. The authors of \cite{moulon13} showed that an iterative use of this methods, adjusted with the cycle length weighting of \cite{EnqvistKO11}, can remove most outlier edges in the graph.
Other approaches \cite{Govindu06,OlssonE11a} are based on random spanning trees, in a RANSAC-like method.

These strategies are computationally demanding and do not scale well with the number of cameras.
% In particular, approaches based on RANSAC suffer from the limitation of increased computational complexity for large-scale datasets.
%and speed is always traded-off with detection accuracy.
For example, \cite{OlssonE11a} reports that outlier removal is the most expensive step (after feature extraction and matching) within the entire SfM pipeline.

% quelli con funzione costo robusta

Recently, a few approaches have been developed to robustly solve the \ARE problem \emph{without} detecting outlier rotations explicitly. 
Techniques in \cite{Hartley11,Chatterjee_2013_ICCV,Wang2013,disco}, together with the approach presented in this paper, come under this category. 

In \cite{Hartley11,Wang2013} a cost function based on the
$\ell_1$ norm is used to average relative rotations, exploiting the fact that
the $\ell_1$ norm is more robust to outliers than the $\ell_2$ norm. 
In \cite{Hartley11} the geodesic (angular) distance is used, while in \cite{Wang2013} the chordal metric is adopted.
The authors of \cite{Wang2013} consider a semidefinite relaxation and use the alternating direction augmented Lagrangian method to minimize the cost function, while
in \cite{Hartley11} each absolute rotation is updated in turn by applying the Weiszfeld algorithm to its neighbours.

In \cite{disco} a truncated quadratic is used as a more robust self-consistency error.
This method uses a discrete Markov random field formulation, combined with a continuous Levenberg-Marquardt refinement.
In addition to relative rotations, vanishing points and information from other sensors are assumed as input.

As observed in \cite{Chatterjee_2013_ICCV}, neither \cite{disco} nor the Weiszfeld algorithm satisfies both the requirements of a computationally efficient and scalable 
robust scheme. On one hand, the Weiszfeld method scales poorly with large datasets, since any change in a given rotation takes a long time to propagate over the entire epipolar graph.
On the other hand, the method in \cite{disco} can handle large-scale problems, but it requires a significant amount of memory.

To overcome these drawbacks, the authors of \cite{Chatterjee_2013_ICCV} proposed a two-stages synchronization scheme that extends the Lie-averaging algorithm in \cite{Govindu04}.
First, the $\ell_1$ norm of a vector that contains both noise and outliers is minimized, exploiting recent work in compressed sensing. Then, this solution is improved
through iteratively reweighted least squares (IRLS). Experiments in \cite{Chatterjee_2013_ICCV} demonstrate that this technique is an efficient solution to the synchronization problem
even for large-scale datasets.

In \cite{Wang2013}, the authors focus on accuracy rather than efficiency, providing theoretical results about exact and stable recovery of rotations.

%----------------------------------------------------------------------------------------
\section{Low-rank and sparse matrix decomposition}
\label{background}

We will show in Section \ref{our_method} that the \ARE problem can be
translated to finding the ``low-rank and sparse decomposition'' of a data matrix which contains a set of noisy and incomplete relative rotations, possibly corrupted by gross errors.

Matrix decompositions have a long history and occurs in the analysis of  complex data. The idea is that decomposing a data matrix into the sum of terms with specific properties makes the understanding easier as it separates information into simpler pieces. In recent years, decompositions imposing constraints on the rank and sparsity of the addends have become very popular thanks to their profitable application in several fields, such as pattern recognition,  machine learning,  and signal processing.

% In particular, low-rank and sparse matrix decompositions allow to recover low-dimensional structures contained into high dimensional data eventually affected also by missing observations, gross errors and diffuse noise. 

Let $\widehat{X}$ be a data matrix,
%whose entries are known (fully or partially)
and suppose that $\widehat{X}$ is known to be the exact or approximate sum of a low-rank term and a sparse term. Low-rank and sparse decompositions address problems of the general form 
\begin{equation}\label{eq:general_form}
\mathcal{F}(\widehat{X}) = \mathcal{F}(L) + S + N
\end{equation}
where $\mathcal{F}$ is a linear operator, $L$ is an unknown low-rank matrix, $S$ is an unknown sparse matrix and $N$ is a diffuse noise. Generally, the sparse term $S$ represents gross errors affecting the measurements (outliers), while the low-rank
part represents some meaningful low-dimensional structure contained into the
data. 
The goal is to recover $L$ (and possibly $S$) under different conditions for $S,N$ and $\mathcal{F}$. 
A survey on this topic is reported in \cite{ZhouYZY14}.

\subsection{Robust Principal Component Analysis}

\begin{figure}[htbp]
\centering
\includegraphics[width=0.5\textwidth]{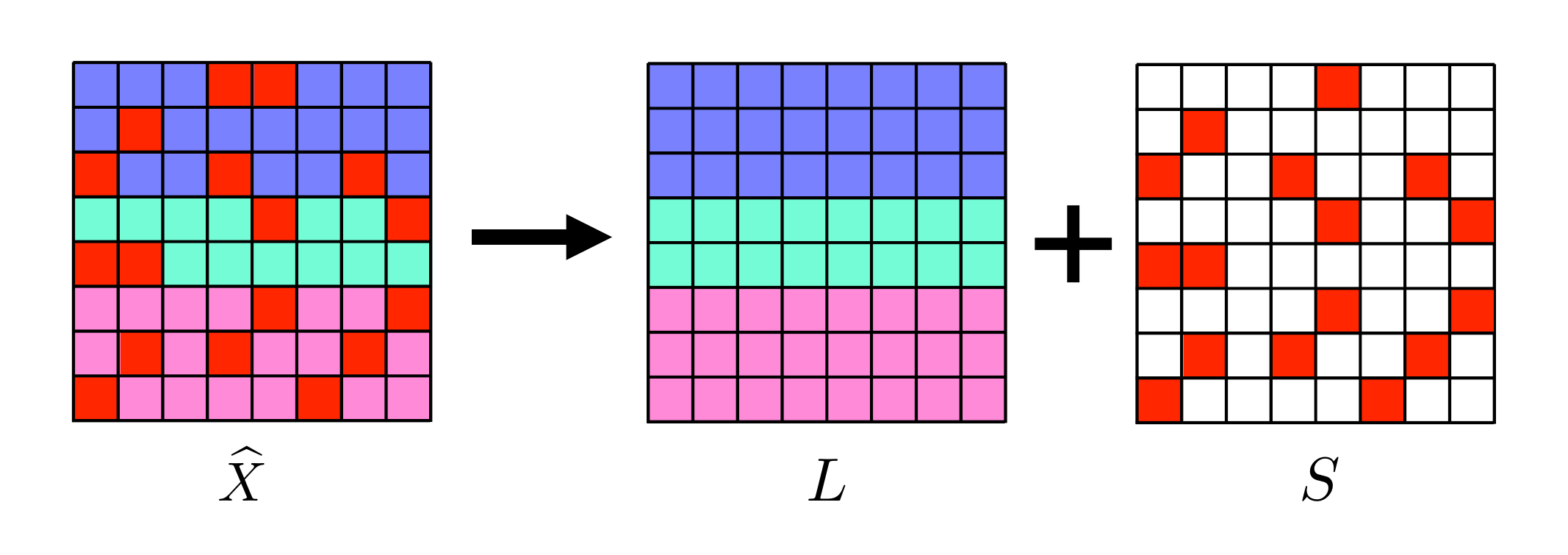}
\caption{Robust Principal Component Analysis. $S$ is the outlier term.}
\label{fig_RPCA}
\end{figure}

An example of low-rank and sparse decomposition is Robust Principal Component Analysis (RPCA) \cite{rpca}. 
The goal is to find the lowest-rank matrix $L$ and the sparsest matrix $S$ such that a given data matrix $\widehat{X}$ can be decomposed as 
\begin{equation}
\widehat{X} = L + S + N
\label{model_RPCA}
\end{equation}
with $N$ a diffuse noise. This is illustrated in Figure \ref{fig_RPCA}. Please observe that such a decomposition is an  instance of the general  problem \eqref{eq:general_form} with $\mathcal{F}$ being the identity operator.

A suitable minimization problem for RPCA is
\begin{equation}
\begin{cases}
\displaystyle \min_{L,S} \left\|L\right\|_{*} + \lambda \left\|{S}\right\|_1 \\
\text{s.t. } \left\| \widehat{X} -L-S\right\|_{F}\leq\epsilon
\end{cases}
\label{eq_RPCA}
\end{equation} 
where $\left\| \cdot \right\|_*$ denotes the nuclear norm, $\left\| \cdot \right\|_F$ denotes the Frobenius norm, $\left\| S \right\|_1$ is the $\ell_1$-norm of $S$ (viewed as a vector), and $\epsilon$, $\lambda$ are given parameters. It is well known from sparse representation theory that minimizing the $\ell_1$-norm promotes sparse vectors \cite{sparse}. Moreover, the nuclear norm is the tightest convex relaxation of the rank function \cite{nuclear_ball}, since it is the \emph{sum} of the singular values of a matrix.
%(while the rank is  the number of its non zero singular values). 
Thus the solution of problem \eqref{eq_RPCA} is expected to recover a blind separation of the lowest-rank component and the sparsest errors contained into the data, i.e. the outliers. 

Theoretical conditions under which such a solution is stable with respect to a diffuse noise $N$ with high probability are studied in \cite{zhou2010stable} and they depend on some incoherence properties of the data matrix and on the sparsity pattern of $S$.

Available algorithms for RPCA include, among others, the Accelerated Proximal Gradient (APG) method \cite{zhou2010stable} and extensions of the Augmented Lagrange Multipliers (ALM) method such as \cite{alm} or the ASALM %method (and its two variants VSALM and PSALM) 
algorithm \cite{tao2011recovering}.  These approaches however involve repeated computation of the SVD (or at least of a partial SVD) of matrices of considerable size which represents the principal bottleneck of current solutions for RPCA.

A faster alternative to RPCA is represented by randomized approximate matrix decomposition \cite{tropp2009finding}. This approach 
proves that the low-rank term $L$ of a decomposition of the form \eqref{model_RPCA} can be well approximated by random projections onto its column space, thus providing a fast approximation of SVD. 

A technique exploiting this paradigm is the \textsc{GoDec} algorithm described in \cite{godec}. This method requires to know approximately both the rank $r$ of the low-rank term $L$ and the cardinality (i.e. the number of non-zero entries) $k$ of the sparse term $S$, and it solves the following minimization problem     
\begin{equation}
\begin{cases}
\displaystyle \min_{L,S} \left\|{ \widehat{X}  - L - S}\right\|_F^2 \\
\text{s.t. rank}(L) \le r, \ \text{ card}(S)\le k.
\end{cases}
\label{eq_GoDec}
\end{equation}
\textsc{GoDec} adopts a \emph{block-coordinate minimization scheme} (a.k.a.~\emph{block relaxation}), i.e. it alternatively forces $L$ to the rank-$r$ approximation of $\widehat{X}-S$, and forces $S$ to the sparse approximation with cardinality $k$ of $\widehat{X}-L$. 

The rank-$r$ projection is computed using Bilateral Random Projections (BRP) instead of SVD thus obtaining a speed up in the computation. The updating of $S$ is obtained via entry-wise hard thresholding, keeping the $k$ largest elements of 
$| \widehat{X}-L |$ only. It can be shown \cite{godec} that the value of the cost function monotonically decreases and converges to a local minimum, while $L$ and 
$S$ linearly converge to local optima.

Estimating the cardinality $k$ of the sparse term might be unreliable in practical applications.
% as one could fall into the same thorny outlier rejection problem which was originally trying to bypass.
In order to avoid this parameter, one  can  consider the following minimization problem instead of \eqref{eq_GoDec}
\begin{equation}
\begin{cases}
\displaystyle \min_{L,S} \frac{1}{2} \left\|{ \widehat{X} - L-S}\right\|_F^2 + \lambda \left\|{S}\right\|_1 \\
\text{s.t. rank}(L) \le r
\end{cases}
\label{L1}
\end{equation}
where $\lambda$ is a regularization parameter which balances the tradeoff between the sparsity of $S$ and the residual error $\left\|{ \widehat{X} - L-S}\right\|_F^2$. 
In this case, the updating of the sparse part is obtained by minimizing the cost function in \eqref{L1}
with respect to $S$, keeping $L$ constant.  Such a problem is known to have an
analytical solution, given by the \emph{soft thresholding} or \emph{shrinkage} operator $\Theta_{\lambda}$ \cite{Beck:2009}
applied to the matrix $\widehat{X}-L$. This operator is defined as follows
\begin{equation}
\Theta_{\lambda} (S) = \text{sign}(S) \cdot \max (0, |S|-\lambda )
\end{equation}
where scalar operations are applied element-wise.

Since this method is at the basis of our development, it is  described in detail in Algorithm \ref{godec}. 

%%More details on this technique can be found in \cite{godec}.

\begin{algorithm}
 \caption{\textsc{GoDec for RPCA}}
 \begin{algorithmic}
 
 \REQUIRE $\widehat{X}$, $r$, $\epsilon$, $\lambda$
 \ENSURE $L$, $S$

\STATE \textbf{Initialize:} $L= \widehat{X} $, $S=0$

\WHILE{$\left\| \widehat{X} -L-S \right\|_F^2 / \left\| \widehat{X} \right\|_F^2 > \epsilon$}
\STATE
\begin{enumerate}
\item  $L \;\; \leftarrow$  rank-$r$ approximation of $ \widehat{X} -S$ via BRP
%\item Assign the rank$-r$ approximation of $ \widehat{X} -S$ to $L$ using BRP
\item $S \;\; \leftarrow \Theta_{\lambda} (\widehat{X}-L) $
%\item $S \;\; \leftarrow  (\widehat{X} -L) \circ \Upsilon$ where 
%$\Upsilon$ is the  nonzero subset of the $k$ largest entries of $| \widehat{X} -L |$
%\item Assign the projection onto $K$ of $ \widehat{X} -L$ to $S$, where $K$ is the nonzero subset of the first $k$ largest entries of $| \widehat{X} -L |$
\end{enumerate}
\ENDWHILE

%\STATE Return $L,S$

\end{algorithmic}
\label{godec}
\end{algorithm}

A principled choice of $\lambda$, which plays a role similar to  an inlier threshold, is derived in \cite{Donoho} in the case of uncorrelated residuals
\begin{equation}
\lambda = \sigma \sqrt{2 \log (m)}
\label{eq_lambda}
\end{equation}
where $m$ is the number of observations and $\sigma$ is an estimate of the noise standard deviation.

%-------------------------------------------------------------------------------------------------
\subsection{Matrix completion}

RPCA assumes that the data
matrix $ \widehat{X} $ is fully available.  However, in
practical scenarios, one has to face the problem of missing data.
Matrix Completion (MC)  \cite{power_MC,exact_MC} is the most natural tool to manage matrices containing unspecified  entries.

\begin{figure}[htbp]
\centering
\includegraphics[width=0.4\textwidth]{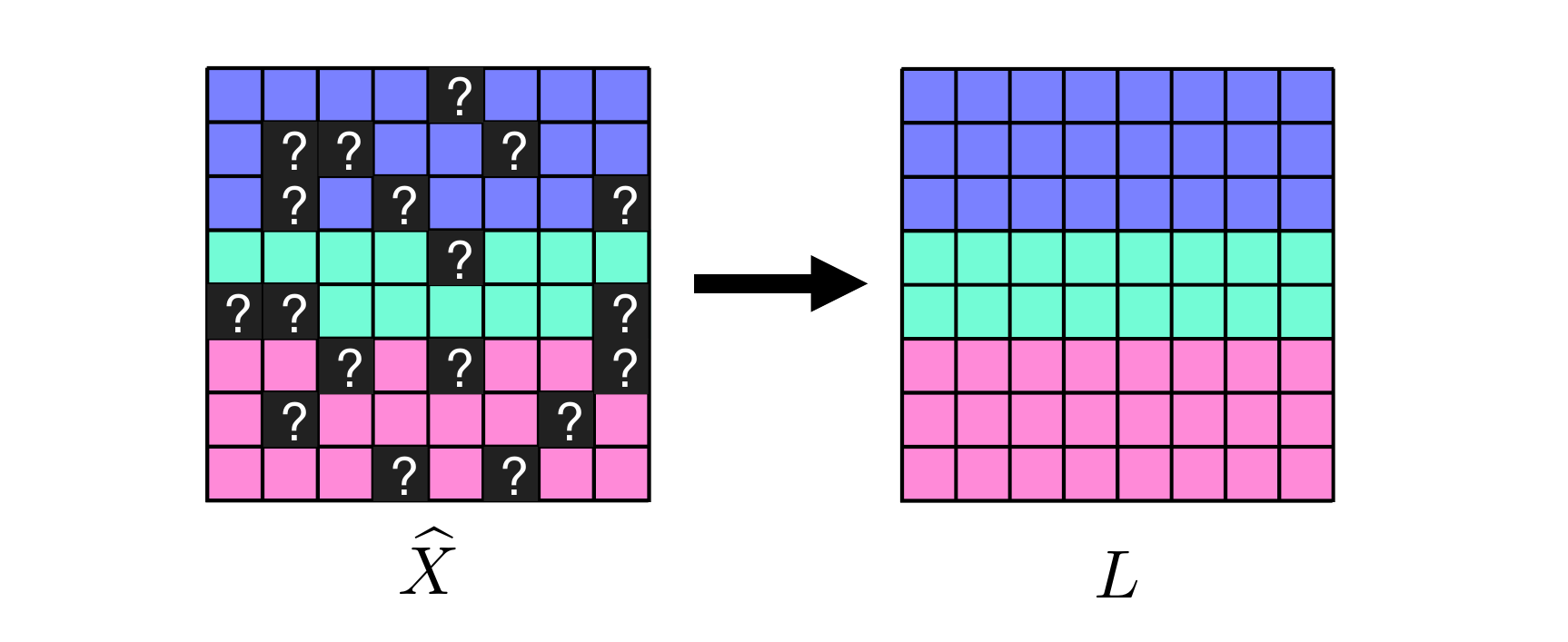}
\caption{$L$ is a low rank completion of $\widehat{X}$.}
\label{fig_MC}
\end{figure}

A \emph{partial} matrix  is a matrix whose entries are specified on a subset of index pairs and unspecified  elsewhere; a \emph{completion} of a partial matrix consist in assigning values to the  unspecified entries. Matrix completion problems are concerned with computing a completion of a partial matrix (if any)  which satisfies some prescribed properties, notably low-rank or positive definiteness. We are concerned here with the low-rank problem, illustrated in Figure \ref{fig_MC}, which can be cast as an instance of the general decomposition \eqref{eq:general_form} with a specific choice of $\mathcal{F}$ and $S=0$, namely
% Actually, the goal of matrix completion techniques is to complete a low-rank data matrix $\widehat{X}$ starting from a random subset of its entries $\mathcal{P}_{\upOmega}( \widehat{X} )$ eventually corrupted with a small amount of noise, as shown in Figure \ref{fig_MC}.
% As a matter of fact, matrix completion is concerned with the decomposition problem 
\begin{equation}
\mathcal{P}_{\upOmega}( \widehat{X} ) = \mathcal{P}_{\upOmega}(L) + N.
\label{eq_MC_original}
\end{equation}
Here $\upOmega$ is a $(0,1)$-matrix  representing the \emph{pattern} of $\widehat{X}$, i.e.  $\upOmega_{ij}=1$ if $\widehat{X}_{ij}$ is specified and $\upOmega_{ij}=0$ otherwise, and $\mathcal{P}_{\upOmega}(X) = \upOmega \circ X$, where $\circ$ is the Hadamard (component-wise) product, with the provision that an unspecified value multiplied by 0 gives 0.

% with $N$ a diffuse noise. In other terms, matrix completion can be seen as an example instance of the decomposition problem \eqref{eq:general_form} with $\mathcal{F} = \mathcal{P}_{\upOmega}$ and $S=0$.  

The MC problem can be solved through nuclear norm minimization  
\begin{equation}
\begin{cases}
\displaystyle \min_{L} \left\|L\right\|_{*}  \\
\text{s.t. } \left\| \mathcal{P}_{\upOmega} ( \widehat{X} - L) \right\|_{F}\leq\epsilon
\end{cases}
\label{eq_MC_nuclear}
\end{equation}
or -- if the rank is known a priori -- by addressing the following optimization problem
\begin{equation}
\begin{cases}
\displaystyle\min_{L} \left\|\mathcal{P}_{\upOmega}( \widehat{X} - L) \right\|_F^2 \\
\text{s.t. rank}(L) \le r.
\end{cases}
\label{eq_matrix_completion}
\end{equation}

Similarly to RPCA, theoretical conditions under which the solution to problem \eqref{eq_MC_nuclear} recovers the full low-rank matrix $L$ with high probability depend on some incoherence properties of the data matrix and on the cardinality and randomness of $\upOmega$ (see \cite{power_MC} for more details).

Conventional solvers for MC include convex solvers such as ALM \cite{alm}, SVT \cite{SVT} and FPCA \cite{FPCA}, and subspace identification solvers such as \textsc{OptSpace} \cite{optspace_paper} and ADMiRA \cite{ADMIRA}. 

Specifically, in the subspace identification problem the goal is to identify the column space of the unknown low-rank term $L$. Clearly, any matrix $L$ of rank up to $r$ admits a factorization of the form $L = ZY^T$ where $Z$ and $Y^T$ are of $r$ columns and $r$ rows respectively. Thus an alternative minimization for the MC problem is
\begin{equation}\label{eq:low_rank_fitting}
\begin{cases}
\displaystyle\min_{L,Z,Y} \frac{1}{2}\left\|ZY^T - L\right\|_{F}^{2} \\
\text{ s.t. } \mathcal{P}_{\upOmega}( \widehat{X} ) = \mathcal{P}_{\upOmega}(L).
\end{cases}
\end{equation}
In particular, \textsc{OptSpace} solves a normalized version of the previous problem, with $Z,Y$ belonging to the Grassmannian manifold, namely the set of all $r$-dimensional subspaces of a Euclidean space, via gradient descent.

The MC problem can also be solved by
 modifying the \textsc{GoDec} Algorithm, as explained in \cite{godec}. The
minimization problem \eqref{eq_matrix_completion} is reformulated by introducing a
sparse term $S$ which approximates $-\mathcal{P}_{\mho} (L)$, where
$\mho$ represents the complementary of $\upOmega$, resulting in the following problem
\begin{equation}
\begin{cases}
\displaystyle\min_{L,S} \left\|{ \mathcal{P}_{\upOmega} ( \widehat{X} ) - L - S }\right\|_F^2 \\
\text{s.t. rank}(L) \le r, \text{ supp}(S) = \mho.
\end{cases}
\label{eq_MC}
\end{equation}
where $\text{supp}(S)$ denotes the support of $S$, i.e.
the $(0,1)$-matrix with $ij$-th entry equal
to 1 if $S_{ij} \neq 0$, and equal to 0  otherwise.
The associated decomposition problem is
\begin{equation}
\mathcal{P}_{\upOmega} ( \widehat{X} ) = L + S + N 
\label{model_MC_godec}
\end{equation}
which is equivalent to \eqref{eq_MC_original} but it does not involve the projection operator $\mathcal{P}_{\upOmega}$ in the right side, thanks to the introduction of the auxiliary variable $S$.
Note that here $S$ does not represent the outliers, but the recovery of
missing entries. In the \textsc{GoDec} algorithm for MC, the updating
of the sparse term is obtained by assigning $\mathcal{P}_{\mho}( \widehat{X} -L) = -\mathcal{P}_{\mho} (L)$ to
$S$. The method is summarized in Algorithm \ref{godec_mc}.

\begin{algorithm}
 \caption{\textsc{GoDec for MC}}
 \begin{algorithmic}
 
 \REQUIRE $ \widehat{X} $, $\upOmega$, $r$, $\epsilon$
 \ENSURE $L$, $S$

\STATE \textbf{Initialize:} $L= \widehat{X} $, $S=0$

\WHILE{$\left\|  \mathcal{P}_{\upOmega} ( \widehat{X} )-L-S \right\|_F^2 / \left\| \mathcal{P}_{\upOmega} ( \widehat{X} ) \right\|_F^2 > \epsilon$}
\STATE
\begin{enumerate}

\item $L \;\; \leftarrow$  rank-$r$ approximation of $ \mathcal{P}_{\upOmega} ( \widehat{X} ) -S$ via BRP
%\item Assign the rank$-r$ approximation of $\widehat{X}-S$ to $L$ using BRP
\item $S  \;\; \leftarrow  -\mathcal{P}_{\mho}(L)$
%\item Assign $\mathcal{P}_{\mho}(\widehat{X}-L) = -\mathcal{P}_{\mho} (L)$ to $S$
\end{enumerate}
\ENDWHILE

%\STATE Return $L,S$

\end{algorithmic}
\label{godec_mc}
\end{algorithm}

%----------------------------------------------------------------------------------------
\subsection{RPCA and MC}
\label{sec:rpcamc}
	
\begin{figure}[htbp]
\centering
\includegraphics[width=0.5\textwidth]{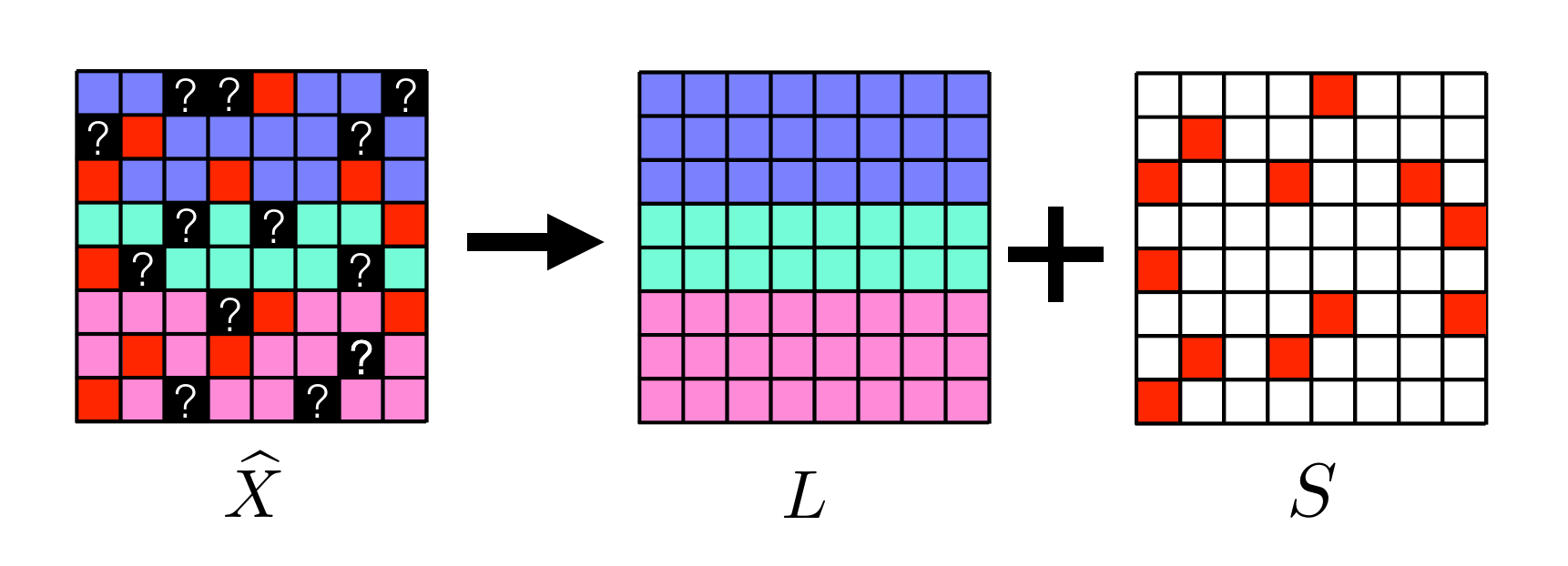}
\caption{Robust Matrix Completion. $L$ is a low rank completion of $\widehat{X} - S$.}
\label{fig_MC_RPCA}
\end{figure}

Although being two instances of the same general formulation \eqref{eq:general_form}, RPCA and MC remain two distinct problems. 
On one hand, RPCA handles the presence of
outlier measurements but it does not deal with missing data, on the other hand MC techniques can fill missing entries, but they are not robust to outliers. 
Addressing these issues simultaneously  is equivalent to solving the following  decomposition problem
\begin{equation}
\mathcal{P}_{\upOmega}( \widehat{X} ) = \mathcal{P}_{\upOmega}(L) +  S  + N
\label{eq_general}
\end{equation}
which aims at recovering the low-rank matrix $L$ starting from an incomplete subset of its entries which are corrupted by both noise and outliers (illustrated in Figure \ref{fig_MC_RPCA}).

This problem
is numerically challenging and poorly studied from a theoretical point of view, as confirmed by the analysis in \cite{ZhouYZY14}. A seminal work is presented in \cite{Waters11sparcs},
where the authors combine a greedy pursuit for updating the sparse term, with
an SVD-based approximation for the low-rank term.
This method requires to know in advance the
cardinality of the sparse term. 
%The only 
Other available approaches are \cite{tao2011recovering}, which reformulates the problem 
% \begin{equation}
% \begin{cases}
% \displaystyle \min_{L,S} \left\|L\right\|_{*} + \lambda \left\|{S}\right\|_1 \\
% \text{s.t. } \left\|\mathcal{P}_{\upOmega}(X)-\mathcal{P}_{\upOmega}(L)-S\right\|_{F}\leq\epsilon
% \end{cases}
% \label{eq_RPCA_MC}
% \end{equation} 
under the scope of the classical ALM, and \cite{balzano2010online,wen2012solving,wang2014practical} which exploit a different formulation in terms of subspace identification in the presence of outliers. In particular, the \textsc{Grasta} algorithm presented in \cite{balzano2010online} minimizes the following cost function 
\begin{equation}\label{eq:grasta}
\begin{cases}
\displaystyle\min_{S,Z,Y}\left\|S\right\|_1  \\
\text{s.t. } \mathcal{P}_{\upOmega}( \widehat{X} )=\mathcal{P}_{\upOmega}(ZY^T)+S, %\,\, Y,Z \in\mathcal{G}(r,n).
\end{cases}
\end{equation}
with $Z,Y$ belonging to the Grasmannian manifold.
%where $\mathcal{G}(r,n)$ represents the $r$-dimensional Grassmannian manifold of $\mathbb{R}^n$. 
%Although \textsc{Grasta}'s formulation is exact (i.e. the partial observations are required to be fitted exactly), experimental results show that the algorithm is stable under small perturbations of noise.

% As we saw previously, the \textsc{GoDec} algorithms, namely Algorithm
% \ref{godec} and \ref{godec_mc}, can solve RPCA and MC, but not simultaneously. 

% As a matter of fact, the two versions of \textsc{GoDec}, namely Algorithm
% \ref{godec} and \ref{godec_mc} are rather
% orthogonal. On one hand Algorithm \ref{godec} handles the presence of
% outlier measurements but it does not deal with missing data, on the other hand Algorithm \ref{godec_mc}  can fill missing entries, but it is not robust to
% outliers. 

We introduce here a  novel variant of \textsc{GoDec}, dubbed 
\textsc{R-GoDec}, which manages at
the same time both the presence of outliers and unspecified  entries in the data
matrix $ \widehat{X} $.
More in detail, the sparse term is expressed as the sum of two terms $S_1$ and $S_2$ having complementary supports:
\begin{itemize}
\item $S_1$ is a sparse matrix with support on $\upOmega$ representing outlier measurements;
\item $S_2$ has support on $\mho$ and it is an approximation of $-\mathcal{P}_{\mho} (L)$, representing completion of missing entries. 
\end{itemize}
This results in the following model
\begin{equation}
\mathcal{P}_{\upOmega} (\widehat{X}) = L + S_1 + S_2 + N
\label{model_S1_S2}
\end{equation} 
which is the natural combination of the RPCA formulation \eqref{model_RPCA} with the MC formulation \eqref{model_MC_godec} associated to the \textsc{GoDec} algorithm.
Equation \eqref{model_S1_S2} reduces to \eqref{model_RPCA} over $\upOmega$, since $S_2$ is zero in $\upOmega$. 
On $\mho$ instead, Equation \eqref{model_S1_S2} turns to $L+S_2+N=0$, since both $S_1$ and $\widehat{X}$ are zero in $\mho$, and thus $S_2$ must coincide with $-L$ (up to noise) as in the case of Problem \eqref{eq_MC}.

The decomposition Problem \eqref{model_S1_S2} is translated into the following minimization
\begin{equation}
\begin{cases}
\displaystyle\min_{L,S_1,S_2} \frac{1}{2} \left\|{ \mathcal{P}_{\upOmega} ( \widehat{X} ) - L - S_1 - S_2 }\right\|_F^2 + \lambda  \left\|{ S_1 } \right\|_{1} \\
\text{s.t. rank}(L) \le r, \\
 \text{supp}(S_1) \subseteq \upOmega,  \\
 \text{supp}(S_2) = \mho
\end{cases}
\label{eq_R_godec}
\end{equation}
which is solved using a block-coordinate minimization scheme that alternates the steps of Algorithm \ref{godec} and Algorithm \ref{godec_mc}.
First, the rank-$r$ projection of $ \mathcal{P}_{\upOmega} ( \widehat{X} )-S_1-S_2$ -- computed through BRP -- is assigned to $L$.
Then, the sparse terms $S_1$ and $S_2$ are updated separately. 
The outlier term $S_1$ is computed by applying the soft-thresholding operator $\Theta_{\lambda}$ to the matrix $\mathcal{P}_{\upOmega}(\widehat{X}-L)$.
As for the completion term, $-\mathcal{P}_{\mho}(L)$ is assigned to $S_2$, according to the  \textsc{GoDec} algorithm for MC. These steps are iterated until convergence.
Our method, called  \textsc{R-GoDec} where ``R'' stands for ``robust'', is summarized in Algorithm \ref{godec_out}.

\begin{algorithm}
 \caption{\textsc{R-GoDec}}
 \begin{algorithmic}
 
 \REQUIRE $\widehat{X}$, $\Omega$, $r$, $\epsilon$, $\lambda$
 \ENSURE $L$, $S_1$, $S_2$

\STATE \textbf{Initialize:} $L=\widehat{X}$, $S_1=0$, $S_2=0$

\WHILE{$\left\| \mathcal{P}_{\upOmega} ( \widehat{X} )-L-S_1-S_2 \right\|_F^2 / \left\| \mathcal{P}_{\upOmega} (\widehat{X}) \right\|_F^2 > \epsilon$}
\STATE
\begin{enumerate}
\item $L\;\; \leftarrow$ rank-$r$ approximation of $ \mathcal{P}_{\upOmega} (\widehat{X} )-S_1-S_2$ via BRP
\item $S_1 \;\; \leftarrow \Theta_{\lambda}(\mathcal{P}_{\upOmega}(\widehat{X}-L))$ 
\item $S_2 \;\; \leftarrow -\mathcal{P}_{\mho}(L)$
\end{enumerate}
\ENDWHILE
%\STATE Return $L,S_1,S_2$
 \end{algorithmic}
 \label{godec_out}
 \end{algorithm}

The proof of convergence follows the same line as in \cite{godec}: in each step the sub-problem over the coordinate block is solved exactly to its optimal solution (modulo the approximation induced  by BRP), hence the objective function is monotonically decreasing (strictly decreasing away from stationary points). In addition,  the constraints are satisfied all the time, hence \textsc{R-GoDec} produces a sequence of objective values that converge to a stationary point of the objective function.

% The objective function is not increasing after an iteration of Algorithm \ref{godec_out} (modulo the approximation induced  by BRP) since each step has this property, as in every block-coordinate minimization scheme.

% As a matter of fact, a formal proof of convergence of the algorithm is out of the scope of this paper. However, the fact that each step does not increase the objective function is a property shared with any block-relaxation technique (modulo the approximation induced  by BRP). 

%----------------------------------------------------------------------------------------
\section{Problem formulation}
\label{formulation}

Let $R_1, \dots, R_n \in SO(3)$ denote $n$  rotations representing the \emph{absolute}  -- i.e expressed in an external coordinate system -- 
angular attitudes of local reference frames.
Let $R_{ij} \in SO(3)$ denote the ideal (noise-free) relative rotation of the pair $(i,j)$, namely the transformation that maps the reference frame represented by $R_i$ in that associated to $R_j$. 
The link between absolute and relative rotations is captured by the \emph{compatibility constraint}
\begin{equation}
R_{ij} = R_i R_j^{\mathsf{T}}.
\label{eq_compatibility}
\end{equation}
% Relative rotations can be seen as measurements for the ratios of  unknown
% elements of $SO(3)$; finding group elements from noisy measurements of their ratios is also
% known as the \emph{synchronization} problem \cite{Giridhar06,Singer2011}.

Let $\widehat{R}_{ij}$ denote an estimate of $R_{ij}$, which can be computed efficiently using standard techniques. In this paper we use the hat accent to denote noisy measurements. 
In the SfM application, relative rotations are obtained by decomposing the essential matrices through SVD, 
which in turn are computed from a collection of matching points across the input images.
In the case of point-set registration, relative rotations are the output of the Iterative Closest Point (ICP) algorithm, which is applied to pairs of point sets. 

The goal of \ARE is to recover the absolute rotations $R_1, \dots, R_n$ starting from the relative rotations measurements $\widehat{R}_{ij}$, thus leaping from two-view to multi-view information.
In practice only a subset of all the relative rotations is available,  due to the lack of overlap between some pairs of images/scans.  However, there is a significant level of redundancy among relative rotations in general datasets, which can be used to distribute the error over all the nodes, avoiding drift in the solution. Let $A$ be the $(0,1)$-matrix that indicates the available measurements: $A_{ij} =1$ if 
$\widehat{R}_{ij}$ is available, $A_{ij} = 0$ otherwise.
Let us consider the graph whose adjacency matrix is $A$: it must consist of a single connected components, in order
to guarantee solvability of the \ARE problem. Consequently, in the minimal case the graph must be a spanning tree over $n$ nodes, which has $n-1$ edges (i.e. relative rotations).

% Let $\mathcal{E} \subseteq \{1,2,\dots,n\} \times \{1,2,\dots,n\}$ denote the set of available pairs, which can be viewed as the  edges of an undirected finite simple graph $\mathcal{G} = (\mathcal{V}, \mathcal{E})$, where vertices in $\mathcal{V}$ correspond to absolute rotations. 
% In practice this graph is far from complete, due to the lack of overlap between some pairs of images/scans.
% However, there is a significant level of redundancy among relative rotations in general datasets, which can be used to distribute the error over all the nodes, avoiding drift in the solution.

The estimated relative rotations are usually corrupted by a diffused noise with small variance, hence they do not satisfy Equation \eqref{eq_compatibility} exactly. 
Thus the goal is to find the absolute rotations
such that $\widehat{R}_{ij} \approx R_i R_j^T$, resulting in the following
minimization problem
\begin{equation}
\min_{R_i \in SO(3)} \sum_{(i,j) \text{ s.t. } A_{ij}=1}  d( \widehat{R}_{ij}, R_i R_j^T)^p
\label{eq_synchronization}
\end{equation}
where $p \ge 1$ and $d(\cdot,\cdot):SO(3) \times SO(3) \mapsto \mathbb{R}^+$ is a bi-invariant metric.
Problem \eqref{eq_synchronization} is analyzed in depth in \cite{averaging} under the name \emph{multiple rotation averaging}.
Distance measures include quaternion, angular (geodesic) and chordal distances:  each metric is related to a particular parametrization of the rotation space, see \cite{Huynh09} for details.
Note that the solution is determined up to a global rotation, affecting the external coordinate system. 
This fact is inherent to the problem and cannot be resolved without external measurements.

In this paper we consider $p=2$ and the chordal metric, that relates to the natural embedding of $SO(3)$ in $ \mathbb{R}^9$, where
rotations are represented by $3 \times 3$ orthogonal matrices with unit determinant. 
Given two rotations $R$ and $S$, their chordal distance is the distance between their
embeddings in $\mathbb{R}^9$, namely $d_{\text{chord}} (R, S) = ||R - S||_F $. 
Accordingly, we address the following problem
\begin{equation}
\min_{R_i \in SO(3)} \sum_{ (i,j) \text{ s.t. } A_{ij}=1}\left\|{\widehat{R}_{ij}-R_i R_j^T}\right\|_F^2
\label{rotation_estimation}
\end{equation}
which is called $\ell_2$-chordal averaging in \cite{averaging}.
This particular choice allows us to cast the \ARE problem in terms of ``low-rank and sparse''
matrix decomposition, as it will be shown in Section \ref{our_method}.

As observed in \cite{Arie12}, Problem \eqref{rotation_estimation} can be reformulated in a useful equivalent form that takes into account all the relative information at once.
 
Let $R$ be the $3n \times 3$ block-matrix containing the absolute rotations and
let $X$ be the $3n \times 3n$ block-matrix containing the pairwise rotations
 \begin{equation}
R = 
\begin{bmatrix}
R_1 \\
R_2 \\
\dots \\
R_n
\end{bmatrix},
\quad
 X = 
 \begin{pmatrix}
 I & R_{12} & \dots & R_{1n} \\
 R_{21} & I &  \dots & R_{2n} \\
 \dots & & & \dots \\
 R_{n1} & R_{n2} & \dots & I
 \end{pmatrix} 
 \label{eq_X}
\end{equation}
where $I$ denotes the $3 \times 3$ identity matrix.
In other words, each block column of $X$ represents all relative rotations with respect to a single rotation.
Clearly $X$ admits the decomposition % \cite{Arie12}
\begin{equation}
X=R R^T
\end{equation}
and hence it satisfies the following properties:
\begin{itemize}
\item $\text{rank}(X) = 3$;
\item $X$ is symmetric and positive semidefinite.
\end{itemize}
%and hence it is symmetric, positive semidefinite and of rank $3$.
Let $\widehat{X}$ be a noisy version of the ideal matrix $X$ containing the observed relative rotations $\widehat{R}_{ij}$, and let $\upOmega$ be the pattern of $\widehat{X}$, namely $\upOmega = (A \otimes  \vone_{3 \times 3} )$,
where $\otimes$ denotes the Kronecker product and $ \vone_{3 \times 3}$ is a matrix of ones.

% In the case of missing data the data matrix $\widehat{X}$ is modified
% by introducing zero blocks in correspondence of missing pairwise rotations, i.e. the only information about the data matrix is given by $\mathcal{P}_{\upOmega} (\widehat{X})$.

Using this notation, the synchronization problem \eqref{rotation_estimation} reduces to minimizing the squared Frobenius norm of the difference between the observed $\widehat{X}$ and the unknown $X$, projected onto the subset of available entries, where $X$ should satisfy the properties mentioned above. 
This results in the following problem
\begin{equation}
\begin{cases}
\displaystyle\min_{X} \left\|{\mathcal{P}_{\upOmega} ( \widehat{X} - X )}\right\|_F^2 \\
\text{s.t. }X = R R^T, \ R \in SO(3)^n
\end{cases}
\label{eq_rank_relaxation}
\end{equation}
which is equivalent to maximize the following cost functions involving the trace operator
\begin{equation}
\displaystyle\max_{R \in SO(3)^n } \text{trace} ( R^{\mathsf{T}} \mathcal{P}_{\upOmega} (\widehat{X}) R ) 
\label{eq_EIG_relaxation}
\end{equation}

\begin{equation}
\begin{cases}
\displaystyle\max_{X} \text{trace} ( \mathcal{P}_{\upOmega} (\widehat{X})^{\mathsf{T}} X ) \\
\text{s.t. }X = R R^T, \ R \in SO(3)^n.
\end{cases}
\label{eq_SDP_relaxation}
\end{equation}

% Problem \eqref{eq_rank_relaxation} is proposed in \cite{isprsarchives-XL-5-63-2014} and it is the basis of our approach, while Problems
% \eqref{eq_EIG_relaxation}-\eqref{eq_SDP_relaxation} are addressed in \cite{Arie12}.
As observed in \cite{averaging}, these are complex
multi-variable non-convex optimization problems, thus a reasonable approach is
to relax some constraints over the variable $X$ to make the computation
tractable. Three examples of relaxations are described in the following sections.

\subsection{Spectral Relaxation}
\label{sec_EIG}

The spectral relaxation (EIG) for rotation synchronization was introduced in \cite{Singer2011} for $SO(2)$ and extended in \cite{Singer2011b,Arie12} to $SO(3)$.
This technique is based on the observation that -- in the absence of noise -- the columns of R are three eigenvectors of $X$ associated to the same eigenvalue, namely $X R = n R$. 
%In the presence of noise the three leading eigenvectors of $\widehat{X}$ can be seen as an estimate of $R$, if the data matrix is complete.
In the case of missing data this generalizes to the following relation
\begin{equation}
\mathcal{P}_{\Omega} (X) R = ( \underbrace{(A \otimes \vone_{3 \times 3})}_{\Omega} \circ X ) R  = (D \otimes I_{3 \times 3} ) R
\end{equation}
where $D \in \mathbb{R}^{n \times n} $ is the degree matrix associated to $A$, i.e. the diagonal matrix such that $D_{ii}$ is the sum of the $i$-th row of $A$.
Thus in the presence of noise the absolute rotations are recovered from the three leading eigenvectors of  $ (D \otimes I_{3 \times 3} )^{-1} \mathcal{P}_{\Omega} (\widehat{X})$.
It is shown in \cite{Arie12} that this method corresponds to solving Problem \eqref{eq_EIG_relaxation} while forcing 
the entire columns of $R$ to be orthonormal, i.e. $R^{\mathsf{T}} R = I$, instead of imposing the orthonormality constraints on each $3 \times 3$ block $R_i$.

% In the case of missing data, the absolute rotations are recovered from the leading eigenvectors
% of $(D \otimes I_{3 \times 3} )^{-1} \widehat{X}$, where $D$ is a $n \times n$ diagonal matrix containing the number of relative rotations linked to node $i$ in its entry $D_{ii}$, i.e. $D_{ii}$ is the number of observations in the $i$-th row of $A$.

%In the presence of noise the absolute rotations are recovered from the least 3 eigenvectors of the matrix
%\begin{equation}
% (D-A)\otimes \vone_{3 \times 3}) \circ \widehat{X}
%\label{eq_laplacian}
%\end{equation}
%where $D \in \mathbb{R}^{n \times n}$ is the degree matrix associated to $A$, i.e. the diagonal matrix such that $D_{ii}$ is the sum of the $i$-th row of $A$.
% The matrix $(D-A)$ is the Laplacian matrix associated to $A$, which gets ``inflated" to a $3 \times 3$-block structure by the Kronecker product with  
%$\vone_{3 \times 3}$ and then is multiplied entry-wise with $\widehat{X}$.

% In order to reduce the effect of outliers, the authors of \cite{ozyesil2013stable} proposed an iterative application of the EIG method 
% in which at each step wrong relative rotations are removed.
% Specifically, a rotation $\widehat{R}_{ij}$ is considered outlier if the consistency error $r_{ij} = || R_iR_j^T - \widehat{R}_{ij} ||_F$
% is sufficiently large, where $R_1 \dots, R_n$ are the current estimates of the absolute rotations.

This formulation can be easily extended to cater for weighted measurements, which translates in letting the entries of $A$ to vary in $[0,1]$, where $0$ still indicates a missing measurement and the other values  reflect the reliability of the pairwise measurements.

% The eigenvalue method can easily incorporate this information, if $A$ is replaced by $W$ in \eqref{eq_laplacian}.
% The case of missing data corresponds to $W = A$, i.e. $w_{ij} = 1$ if $\widehat{R}_{ij}$ is available and $w_{ij} = 0$ otherwise.
% %The idea of using weights is also suggested in \cite{Wang2013}.

This fact allows a straightforward robust enhancement via 
Iteratively Reweighted Least Squares (IRLS).
Specifically, at each step
the weights $a_{ij}$ of the relative rotations are updated based on the residual errors $r_{ij} = || R_iR_j^T - \widehat{R}_{ij} ||_F$, where $R_1 \dots, R_n$ are the current estimates of the absolute rotations, and this is iterated until convergence.
In our experiments we used the Cauchy weight function \cite{Holland77}, namely $a_{ij} = 1 / (1 + (r_{ij} / c) ^2) $, using the default value for the tuning constant $c$.  This method will be referred to as  EIG-IRLS.

\subsection{Semidefinite Relaxation}

In the semidefinite relaxation the optimization variable $X$ is constrained to be symmetric positive semidefinite, and covered by identity blocks along its diagonal, while the remaining properties on $X$ are not enforced. 
In this way Problem \eqref{eq_SDP_relaxation} becomes a semidefinite program (SDP), yielding a tighter relaxation compared to the eigenvalue method.
More details about this technique can be found in \cite{Arie12}.

\subsection{Rank Relaxation}

Another possibility is the rank relaxation introduced in \cite{isprsarchives-XL-5-63-2014}, where the matrix $X$ is enforced to have rank (at most) $3$. 
As a consequence Problem \eqref{eq_rank_relaxation} reduces to a MC problem (namely Problem \eqref{eq_matrix_completion}) which can be solved by using standard MC solvers, such as the \textsc{OptSpace} algorithm \cite{optspace_paper}.

We observe that it is possible \emph{in theory} to recover the absolute rotations by means of matrix completion, since the minimum number of entries required to complete $\widehat{X}$ coincides with the minimum number of rotations necessary to solve the \ARE problem.
Indeed, the matrix $X$ defined in \eqref{eq_X} depends on $9(2n-1)$ parameters (in general, a $n_1 \times n_2$ matrix of rank $r$ has $(n_1 + n_2 - r ) r$ degrees of freedom).
The minimum number of relative rotations necessary to solve the \ARE problem is $n-1$; in addition to this, we have $n$ rotations equal to the identity matrix 
(i.e. $R_{ii} = I$ for $i=1, \dots, n$) resulting in $9 (2n-1)$ entries.

%----------------------------------------------------------------------------------------
\section{Rotation Synchronization
 via Low-rank and Spar\-se Matrix Decomposition}
\label{our_method}

In this section we cast the rotation synchronization problem as a ``low-rank and sparse matrix decomposition'',  paving 
the way to the application of general matrix decomposition techniques
in structure from motion and multiple point-set registration.
Besides this, we adapt Algorithm \ref{godec_out} in order to fit the needs of rotation synchronization. 

As observed in the previous section, the rotation synchronization problem can be expressed in terms of MC, if the rank relaxation is adopted. Indeed, the ideal matrix $X$ is known to be low-rank, i.e. $\text{rank}(X) = 3 << 3n$, and there are missing data, since not all the relative rotations are specified. 
In addition to this, some of the measures are outliers. Considering the rank relaxation of \eqref{eq_rank_relaxation} with an outlier term that brings in resilience to rogue measures, it results in the following minimization problem
\begin{equation}
\begin{cases}
\displaystyle\min_{X} \left\|{\mathcal{P}_{\upOmega} ( \widehat{X} - L  - S) }\right\|_F^2 
\\
\text{s.t. } \text{rank}(L) \leq 3,\\
 \text{supp}(S) \subseteq \upOmega, \ S \text{ is sparse in } \upOmega
\end{cases}
\end{equation}
This corresponds to the model \eqref{eq_general}, namely
$
\mathcal{P}_{\upOmega}( \widehat{X} ) = \mathcal{P}_{\upOmega}(L) +  S  + N
$, 
which is a low rank and sparse decomposition problem with unspecified entries and outliers.
Alternatively, the equivalent formulation \eqref{model_S1_S2} can be used, namely $\mathcal{P}_{\upOmega} (\widehat{X}) = L + S_1 + S_2 + N$.

It is worth noting that here the outliers are intrinsically included in the cost
function. With respect to non robust solutions that rely on a preliminary
outlier rejection step, this approach has the great advantage of being
\emph{intrinsically} resilient against outliers.

% In other terms, referring to the SfM terminology, the epipolar graph is not complete.
% This corresponds to representing the \ARE problem through model \eqref{eq_MC_original} or \eqref{model_MC_godec}.
% As explained in Section \ref{background}, this formulation can handle a low level of noise among the input data, 
% but it is highly non-robust. 
% In other words, it it can efficiently synchronize noisy inliers, which explain well the absolute rotations, but it can give incorrect results in presence of even a single outlier rotation.
% This limitation affects all the methods related to Problem \eqref{eq_rank_relaxation} and to its equivalent formulations \eqref{eq_EIG_relaxation} -  \eqref{eq_SDP_relaxation}. 
% To overcome this drawback the general decomposition problem \eqref{eq_general} is required in order to cope with outliers, noise and missing rotations at the same time.
% While a low-level of noise with small variance is diffused among all the relative rotations, gross errors are sparse over the viewgraph, thus 
% Equation \eqref{eq_general} is a suitable way to model the rotation synchronization problem. 

Thus the absolute rotations can be recovered through methods addressing RPCA and MC simultaneously, such as the \textsc{Grasta} Algorithm \cite{balzano2010online} or  the \textsc{R-GoDec} Algorithm, as described in Section \ref{sec:rpcamc}.
 
In the \ARE problem,  however, the data matrix $\widehat{X}$ has a block structure, being composed of 3D rotations, and this should be reflected by the sparse term which represents the outliers.
This is taken into account by modifying Algorithm \ref{godec_out} in order to enforce a block-structure in $S_1$.
Specifically,  the $\ell_1$-norm in \eqref{eq_R_godec} is substituted with the mixed $\ell_{2,1}$-norm which promotes group sparsity. 

Accordingly, we address the following problem
\begin{equation}
\begin{cases}
\displaystyle\min_{L,S_1,S_2} \frac{1}{2} \left\|{ \mathcal{P}_{\upOmega} ( \widehat{X} ) - L - S_1 - S_2 }\right\|_F^2 + \lambda  \left\|{ S_1 } \right\|_{2,1} \\
\text{s.t. rank}(L) \le 3, \\
 \text{supp}(S_1) \subseteq \upOmega,  \\
 \text{supp}(S_2) = \mho
\end{cases}
\label{mixed_godec}
\end{equation}
where the mixed $\ell_{2,1}$-norm of a $3n \times 3n$ matrix $S$ is defined as the sum of the Frobenius norm of each $3 \times 3$ block $S_{ij}$
\begin{equation}
\left\|{ S } \right\|_{2,1} = \sum_{i,j = 1}^n \left\|{ S_{ij} } \right\|_F^2 .
\end{equation}

% Note that Problem \eqref{mixed_godec} can be viewed as a robust version of the original rotation synchronization problem \eqref{eq_rank_relaxation} (up to the rank relaxation),
% where robustness is introduced through $\ell_{2,1}$-regularization and the additional variable $S_2$ is introduced to handle missing data.

The minimum of the cost function with respect to $S_1$ keeping the other variables constant has a closed-form expression, given by the \emph{generalized soft-thresholding} (or shr\-inkage) operator $\Theta^{2,1}_{\lambda}$ applied to the matrix $\mathcal{P}_{\upOmega}(\widehat{X}-L)$ \cite{Kowalski09}. Such an operator takes a $3n \times 3n$ matrix $S$ as input and  on each $3 \times 3$ block $S_{ij}$ it computes
\begin{equation}
\Theta^{2,1}_{\lambda} ( S_{ij} ) = S \cdot \max(1- \frac{ \lambda }{  || S_{ij} ||_F } ,0)
\end{equation}
where scalar operations are applied element-wise.
In this way the selected blocks are the ones with the biggest Frobenius norms.
Accordingly, Step 2 of Algorithm \ref{godec_out} is modified as follows
$$
S_1 \;\; \leftarrow \Theta^{2,1}_{\lambda} ( \mathcal{P}_{\upOmega}(\widehat{X}-L) ).
$$

Once the optimal $L$ is found, we proceed as follows to compute the
absolute rotations. Since the solution of rotation synchronization is defined up to a global rotation,
any block-column of $L$ can be used as an estimate of $R$. Due to the rank
relaxation, each $3 \times 3$ block is not guaranteed to belong to $SO(3)$, thus
we find the nearest rotation matrix (in the Frobenius norm sense) by using SVD
\cite{ke75}.

As for the optimal $S_1$, it can be used to identify the outliers, since rogue relative rotations correspond to nonzero blocks in $S_1$.
Although the absolute rotations computed by our algorithm are intrinsically
insensitive to outliers, it might be beneficial for the subsequent steps (e.g.
computing translations in SfM) to single out bad relative rotations from the
data matrix $\widehat{X}$, which are indicators that the whole relative motion (including translation)
is probably faulty.

%\subsection{Generalization to SO(d)}

% ---------------------------------------------------------------------------------------------------

\section{Experiments}
\label{experiments}

We evaluate our solution on both synthetic and real scenarios in terms of accuracy, execution cost and robustness to outliers. 
All the experiments are performed in \textsc{Matlab} on a dual-core MacBook Air with i5 1.3GHz processor.

We compare  \textsc{R-GoDec} to
several techniques from the state of the art.  We consider the spectral relaxation (EIG) \cite{Arie12}, the semidefinite relaxation (SDP) \cite{Arie12}, the rank relaxation (\textsc{OptSpace}) \cite{optspace_paper}, the Weiszfeld algorithm \cite{Hartley11}, the L1-IRLS algorithm \cite{Chatterjee_2013_ICCV}, and the LUD algorithm \cite{Wang2013}. 
We also include in the comparison two ``in-house'' competitors, namely
the EIG-IRLS method described in Section \ref{sec_EIG} 
%(which is similar in  spirit to  \cite{ozyesil2013stable} ) 
and the \textsc{Grasta} method, obtained by plugging \textsc{Grasta} in our framework in place of \textsc{R-GoDec}\footnote{With a little abuse of notation, we will use "\textsc{R-GoDec}" and "\textsc{Grasta}" to denote both the low-rank and sparse decomposition algorithms and the rotation synchronization method that uses them.}.

The code of LUD has been provided by the authors of \cite{Wang2013}, the codes of \textsc{Grasta}, \textsc{OptSpace} and L1-IRLS are available on the web, while in the other cases we used our implementation.
The SeDuMi toolbox \cite{sedumi} has been used to solve the semidefinite program associated to the SDP method.

In order to compare estimated and ground-truth absolute rotations we employ $\ell_1$ single averaging.
Specifically, if $\widehat{R}_1, \dots, \widehat{R}_n$ are estimates of the theoretical absolute rotations $R_1, \dots, R_n$, then the optimal $S \in SO(3)$ that align them into a common reference system is the single mean of the set $\{R_i \widehat{R}_i^T, i=1, \dots, n \}$, and it can be computed e.g. by using \cite{Hartley11}.
Then we use the angular distance to evaluate the accuracy of rotation recovery. 
The angular (or geodesic)
distance between two rotations $A$ and $B$
is the angle of the rotation $BA^T$ (in the angle-axis representation) so
chosen to lie in the range $[0, 180^{\circ}]$, namely
$
d_{\angle} (A,B) = d_{\angle} (BA^T,I) = 1/\sqrt{2} \left\| \log(BA^T)
\right\|_2.
$
Other distances in $SO(3)$ can be considered with comparable results.

\subsection{Simulated Data}

In our simulations we consider $n$ rotation matrices sampled from random Euler angles, representing ground truth absolute orientations.
A fraction of the pairwise rotations is drawn uniformly from $SO(3)$, simulating outliers.
The remaining pairwise rotations are either unspecified or corrupted by multiplicative noise
$
\widehat{R}_{ij} = R_{ij} N_{ij}
$
where $N_{ij} \in SO(3)$ has angle between $1^{\circ}$ and $10^{\circ}$, and axis uniformly distributed over the unit sphere, thus representing a small perturbation of the identity matrix. Considering the first order approximation of rotations, this corresponds to additive noise.
The pattern of missing rotations is sampled uniformly, with the
constraint that the underlying graph remains connected.
This general framework can represent both structure from motion and point-set registration.
All the results are averaged over $20$ trials.

\begin{figure*}[htbp]
\centering
\includegraphics[width=0.4\linewidth]{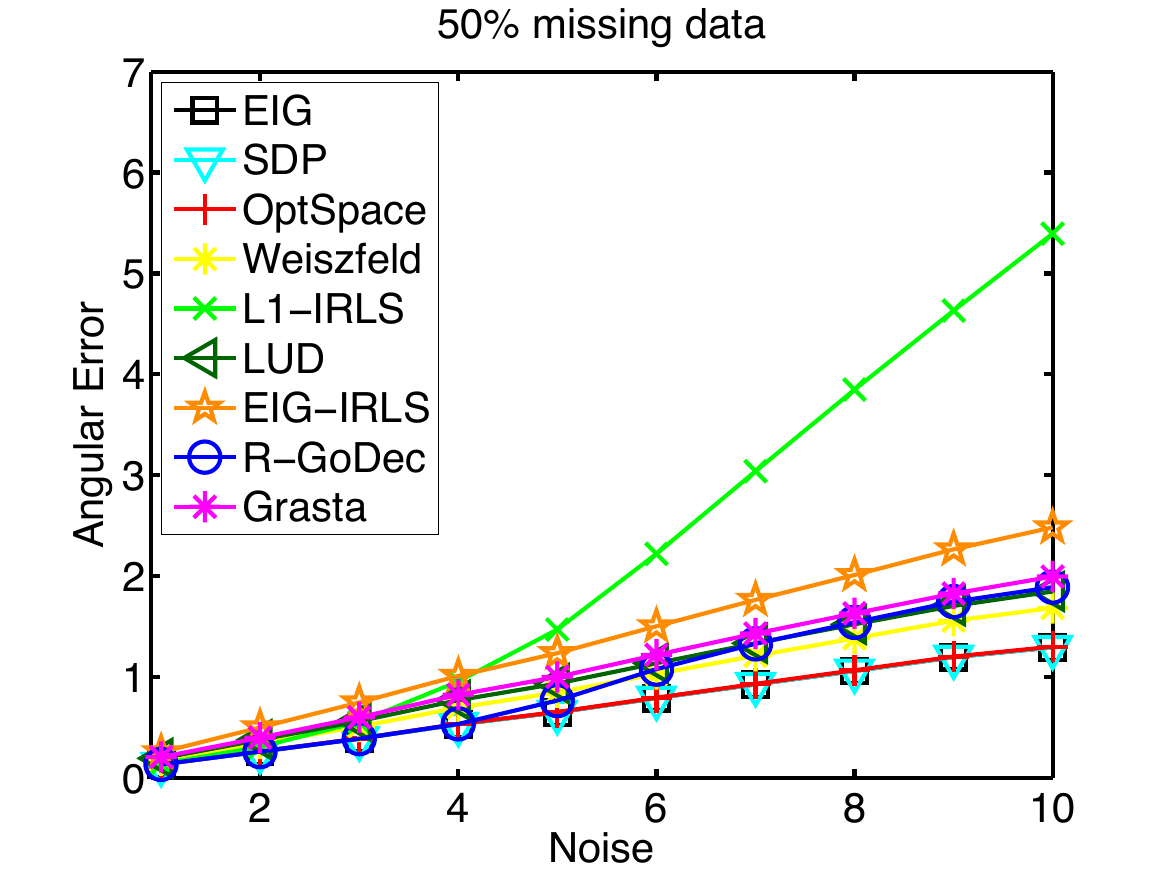}\quad
\includegraphics[width=0.4\linewidth]{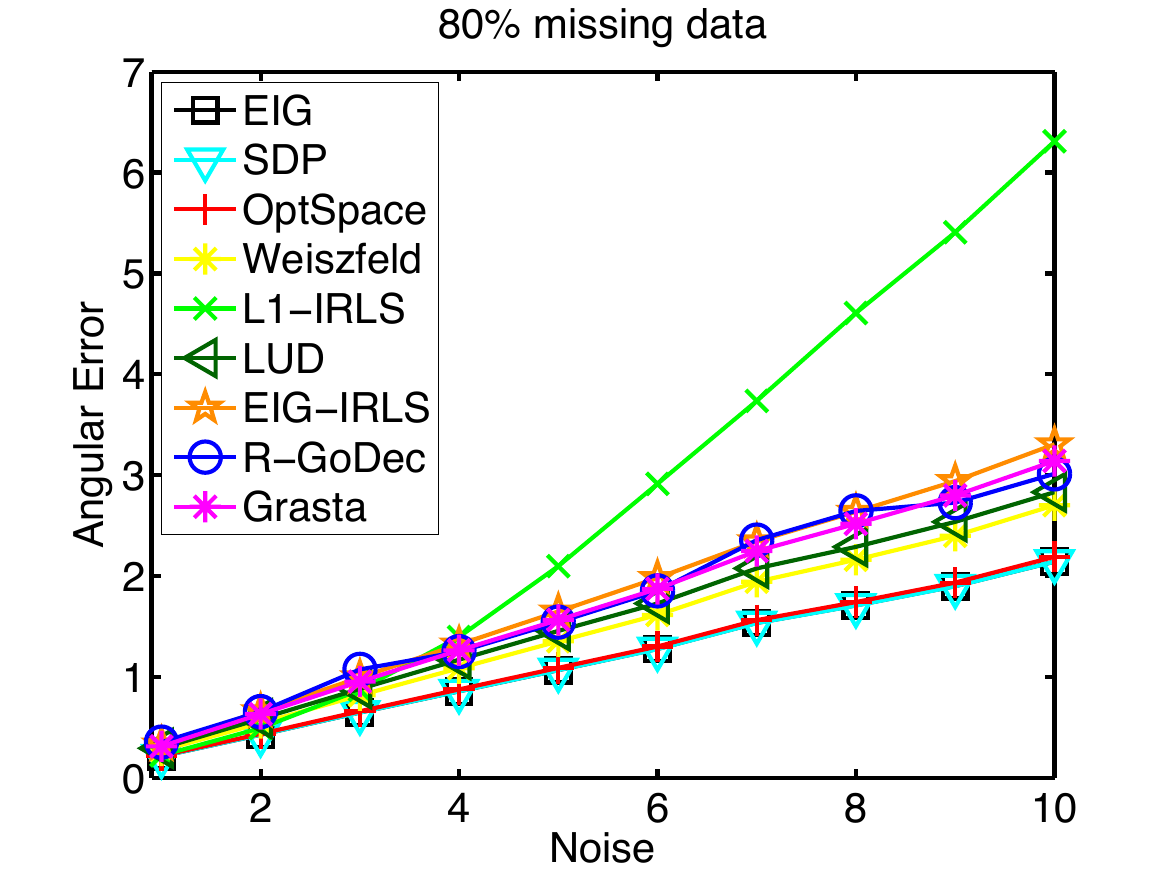}
\caption{Mean angular errors [degrees]  as a function of the noise level, for two different
  percentages of missing relative rotations. Outliers are not introduced in this experiment.}
\label{fig_noise}
\end{figure*}

Figure \ref{fig_noise} shows the behaviour of the aforementioned methods in the presence of noise among the input rotations for two different percentages of missing pairs, with $n=100$; outliers are not introduced in this experiment. As expected the lowest error 
is achieved by EIG together with SDP and \textsc{OptSpace}. On the contrary, all the robust methods
yield worse results, since they essentially trade robustness for statistical efficiency.  
The bad behaviour of L1-IRLS deserves a special discussion. 
This approach works on vectors on the tangent space rather than matrices, and it is similar in principle to our approach since it uses the $\ell_1$-norm as a sparsity promoter.
However, the $\ell_1$-norm of a vector which contains  noise \emph{and} outliers is minimized, yielding an erroneous handling of noise, which is not sparse.

% since noise is not sparse. 

\begin{figure*}[htbp]
\centering
\includegraphics[width=0.4\linewidth]{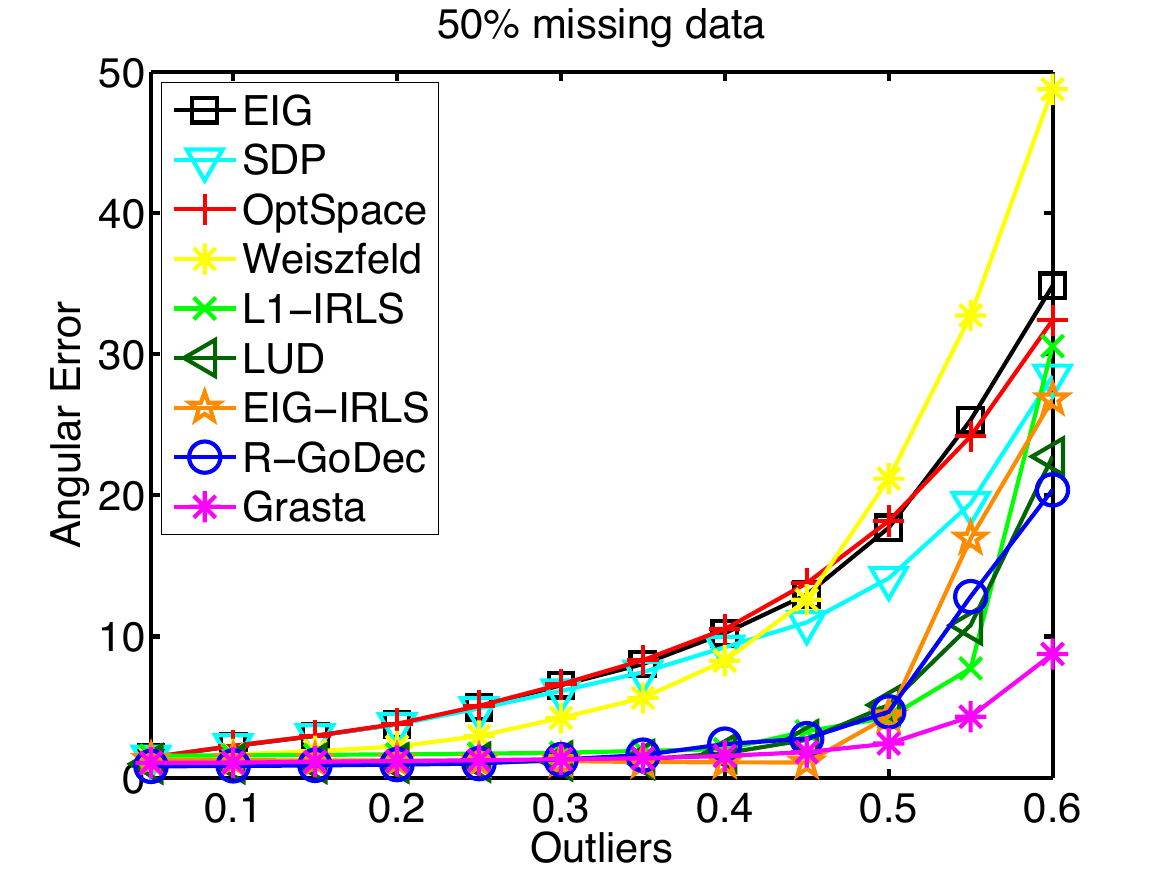}\quad
\includegraphics[width=0.4\linewidth]{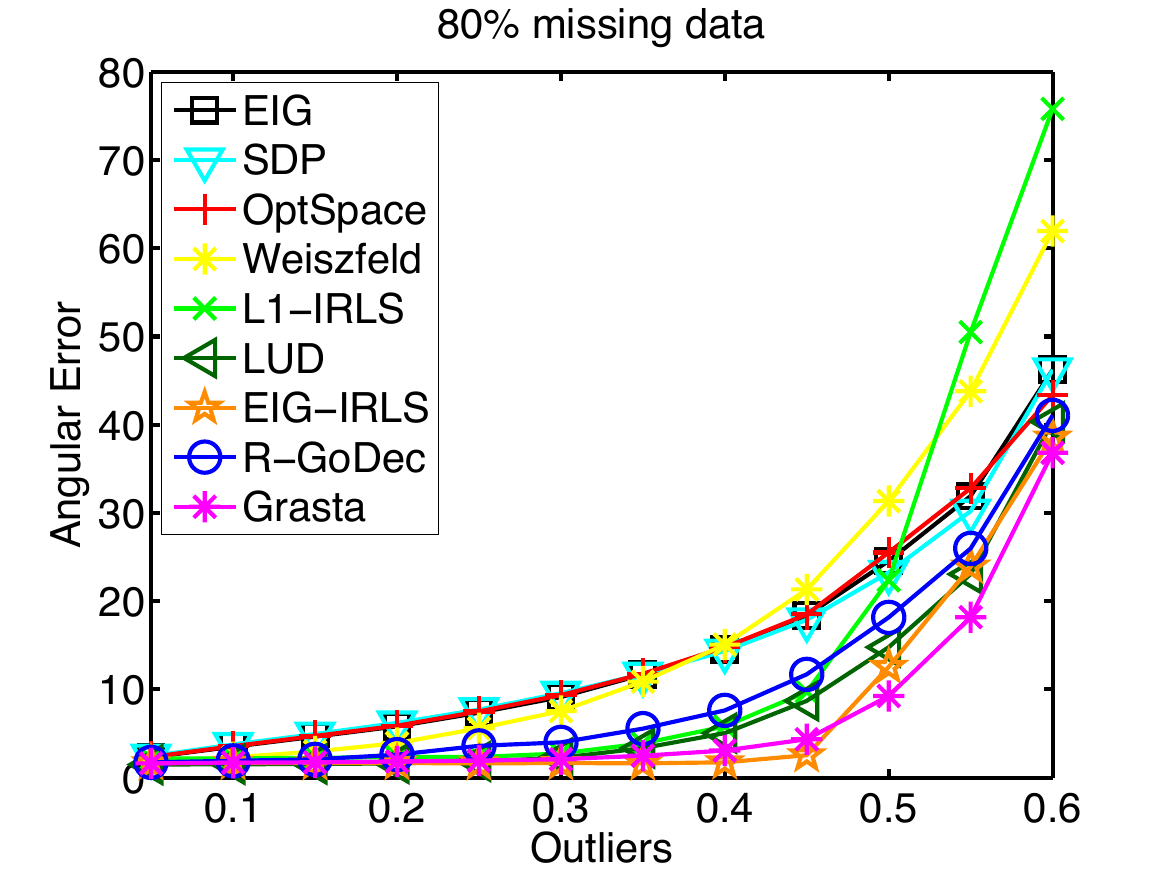}
\caption{Mean angular errors [degrees] as a function of the fraction of outliers, for two different
  percentages of missing relative rotations. A fixed level of noise is applied to the inlier rotations in this experiment.}
\label{fig_outliers}
\end{figure*}

Figure \ref{fig_outliers} shows the angular errors of all the analysed methods as a function of the proportion of outliers, for two different percentages of missing data, with $n=100$. The fraction of wrong rotations is referred to the number of available relative rotations, not to the total number of pairs. 
In this experiments all the inlier rotations are corrupted by a fixed level of noise ($5^{\circ}$).
If the percentage of unspecified relative rotations is $50\%$ and outliers do not exceed inliers, then the error of \textsc{R-GoDec} remains almost constant, showing no sensitivity to outliers. 
%This might suggest that \textsc{R-GoDec} has an empirical $0.5$ breakdown point.
The same happens for \textsc{Grasta, LUD, L1-IRLS} and \textsc{EIG-IRLS}. 
On the contrary, EIG, SDP and \textsc{OptSpace} are non robust to outliers, as already observed in the previous sections.
As for the Weiszfeld algorithm, its performances places it at in middle between robust and non-robust solutions. 
Specifically, it shows good resilience to outlier rotations when they are below $30\%$, then the error starts to grow up, yielding a behaviour similar  to non-robust approaches. 
When the data matrix is highly incomplete (the case of $80\%$ of missing data), the difference between robust and non-robust solutions becomes smaller, however results are qualitatively similar to the previous case.

\begin{figure*}[htbp]
\centering
\includegraphics[width=0.4\linewidth]{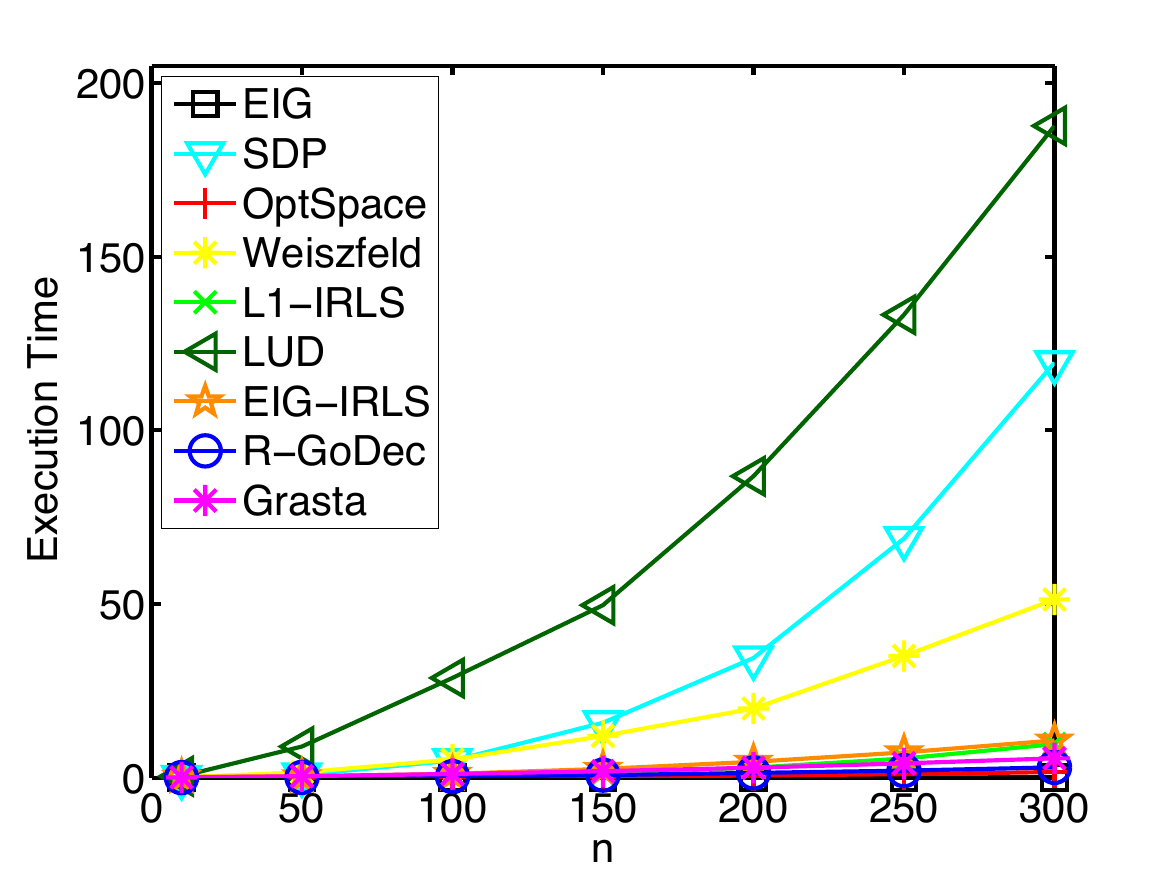}\quad
\includegraphics[width=0.4\linewidth]{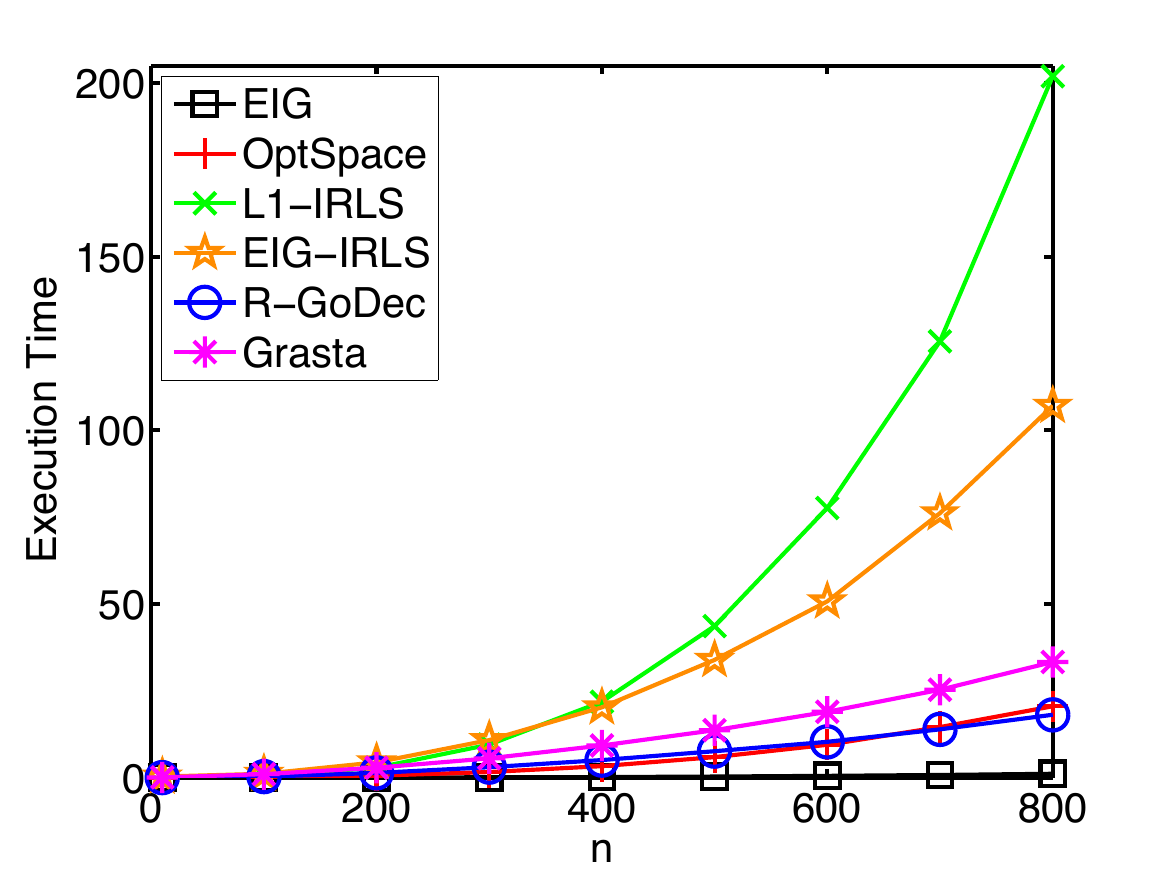}
\caption{Execution times [seconds] as a
  function of the number of absolute rotations. The parameters defining the stopping criterion (maximum number of iterations and tolerance) are the same for all the methods. In the right figure only the fastest solutions are represented, using a wider range in the x-axis.}
\label{fig_time}
\end{figure*}

We conclude this analysis by discussing the performances of rotation synchronization in terms
of computational cost. Figure \ref{fig_time} reports the running time of the analysed methods as a function of $n$. 
Due to computational limitations, in the left sub-figure -- where all the methods are considered -- $n$ reaches $300$, while in the right sub-figure -- where only the fastest methods are reported -- $n$ reaches $800$.
In this experiment we consider a fixed level of missing pairs ($50\%$), noise ($5^{\circ}$) and outliers ($20 \%$).
Figure \ref{fig_time} shows that LUD, SDP and Weiszfeld qualify as the slowest algorithms, while the other ones are significantly faster. 
In particular, the EIG method  is the fastest solution to the \ARE problem,  but it is not robust.
Among all the robust methods, \textsc{R-GoDec} achieves the lowest execution time, outperforming both L1-IRLS and EIG-IRLS. The execution time  of \textsc{Grasta} is only slightly higher  than  \textsc{R-GoDec}.
Note that \textsc{R-GoDec} achieves the same computational cost as \textsc{OptSpace}, which performs matrix completion only, whereas our algorithm performs MC and RPCA simultaneously.

% With respect to \textsc{EIG}, \textsc{R-GoDec}
% buys robustness at a very little computational cost.

With respect to L1-IRLS, which is the leading solution in the context of robust rotation averaging, \textsc{R-GoDec} achieves comparable results as for robustness to outliers, it is more accurate when outliers are not present, and it is faster. 

% The same holds for \textsc{Grasta}, which is tightly related to our novel formulation of the \ARE problem.

% Our method is comparable to spectral decomposition and \textsc{OptSpace} algorithms, which
% however are not robust, and significantly faster than
% semidefinite programming and Weiszfeld algorithms. Considering that \textsc{EIG} is the fastest solution to
% the \ARE problem known in the literature, one can see how \textsc{R-GoDec}
% buys robustness at a very little computational cost.
% The reason why Weiszfeld curve is not so regular in Figure \ref{time} is as follows.
% As for the initialization, Weiszfeld algorithm propagates the compatibility constraint along a random spanning tree, starting from the node with the maximum number of incident edges. 
% Thus, the number of iterations required to yield convergence is dependent on the accuracy of such initial guess. 

%----------------------------------------------------------------------------------

\subsection{Real data}

We apply our algorithm to the structure from motion problem, considering both benchmark \cite{real_data} and irregular large-scale image collections \cite{Snavely14}. We used formula \eqref{eq_lambda} with $\sigma = 0.02$ to define the regularization parameter $\lambda$.

In the first case, ground-truth absolute rotations are available, which we used to evaluate the performances of the analysed methods.
We follow a common SfM pipeline to obtain estimates of relative rotations. 
First, reliable matching points across the input images are computed by using SIFT key-points. Then, each essential matrix is computed in a RANSAC scheme, and it is factorized to obtain a unique $\widehat{R}_{ij}$, which is considered missing if insufficient inlier correspondences are found. The relative rotations are then improved by applying Bundle adjustment to pairs of cameras.

In the second case, ground-truth rotations are not available, thus we use the output of \textsc{Bundler} \cite{Snavely06} as reference solution. We used the estimates of the relative rotations provided by
the authors of \cite{Snavely14} together with the images.

\begin{figure*}[htbp]
\centering
\includegraphics[width=0.3\linewidth]{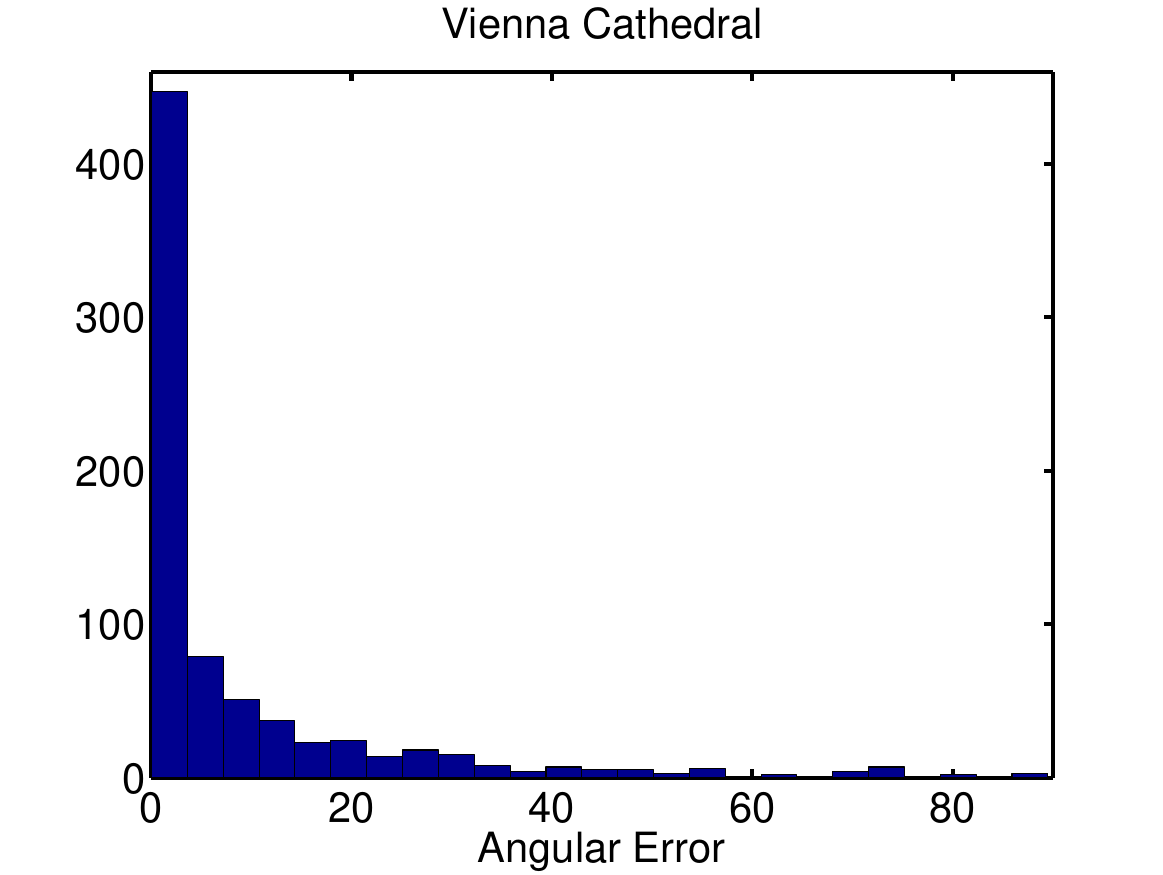}
\includegraphics[width=0.3\linewidth]{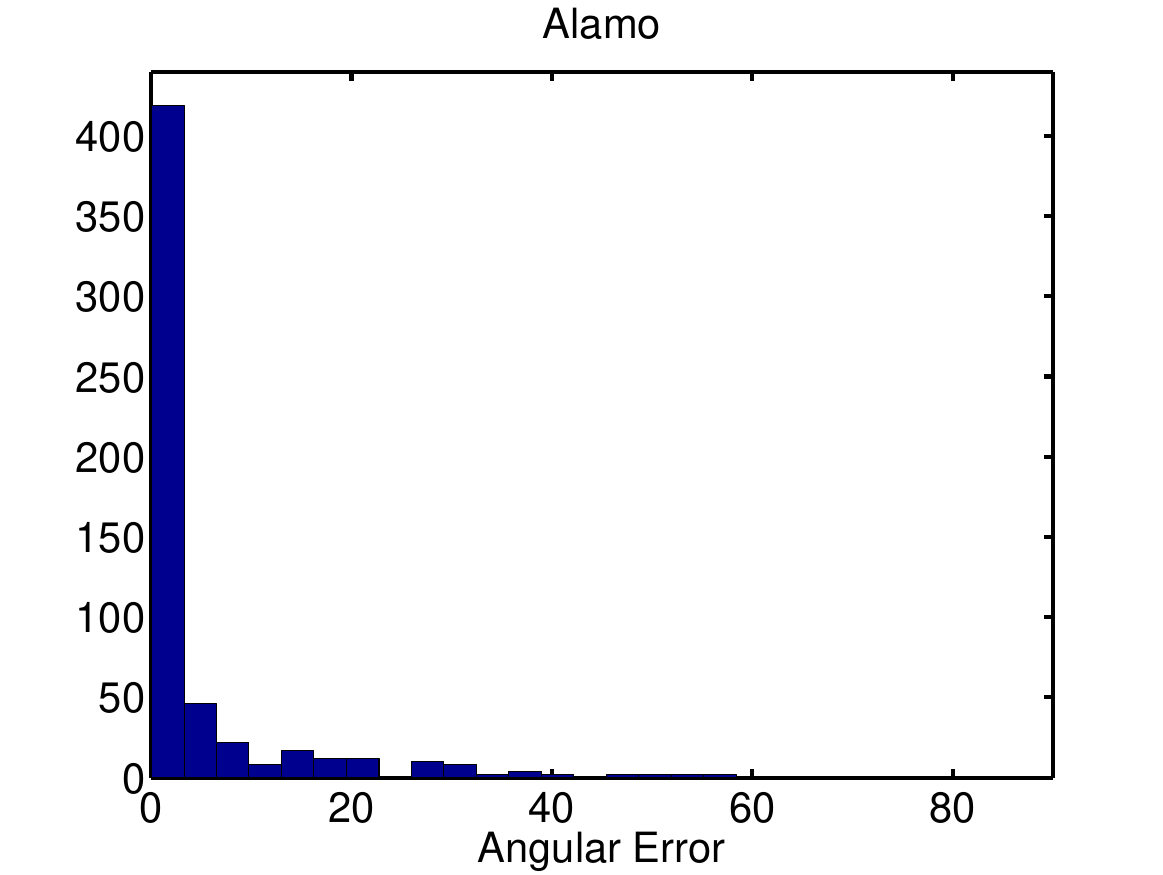}
\includegraphics[width=0.3\linewidth]{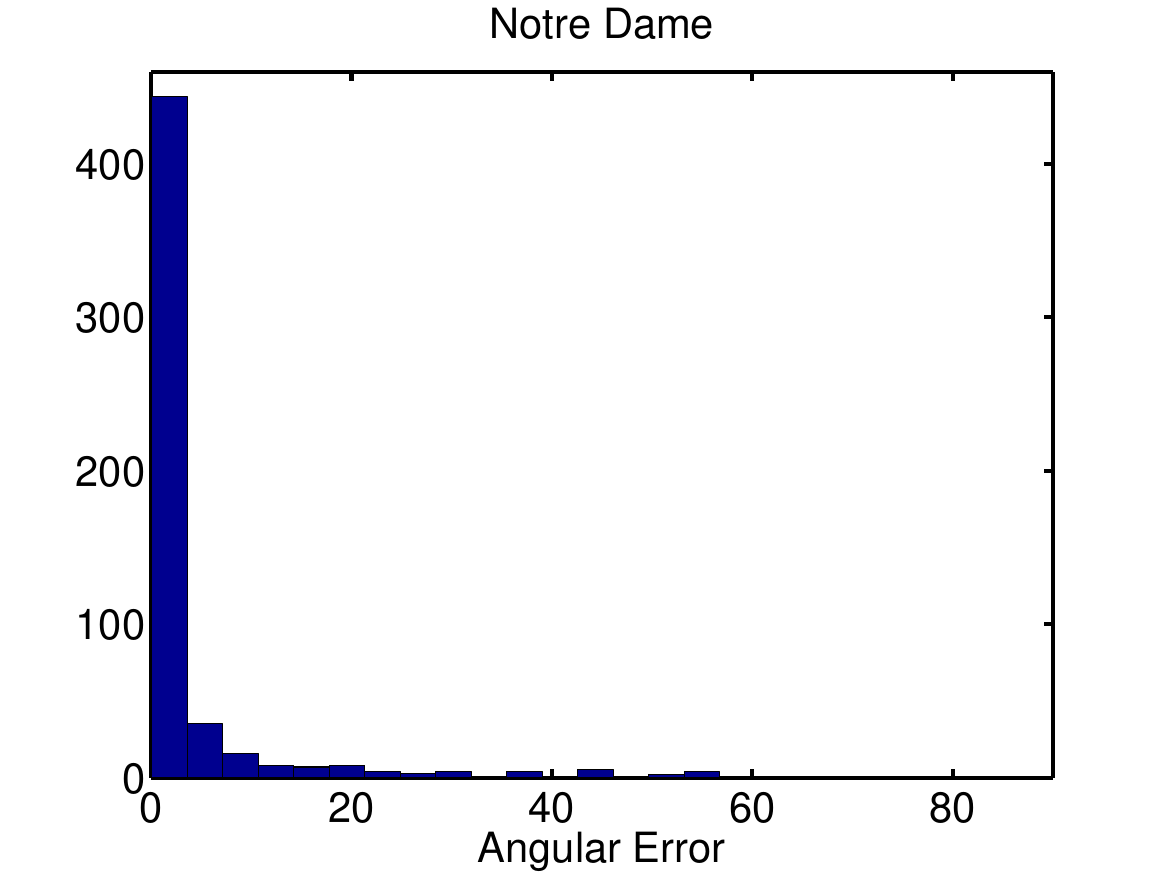} \\
\includegraphics[width=0.3\linewidth]{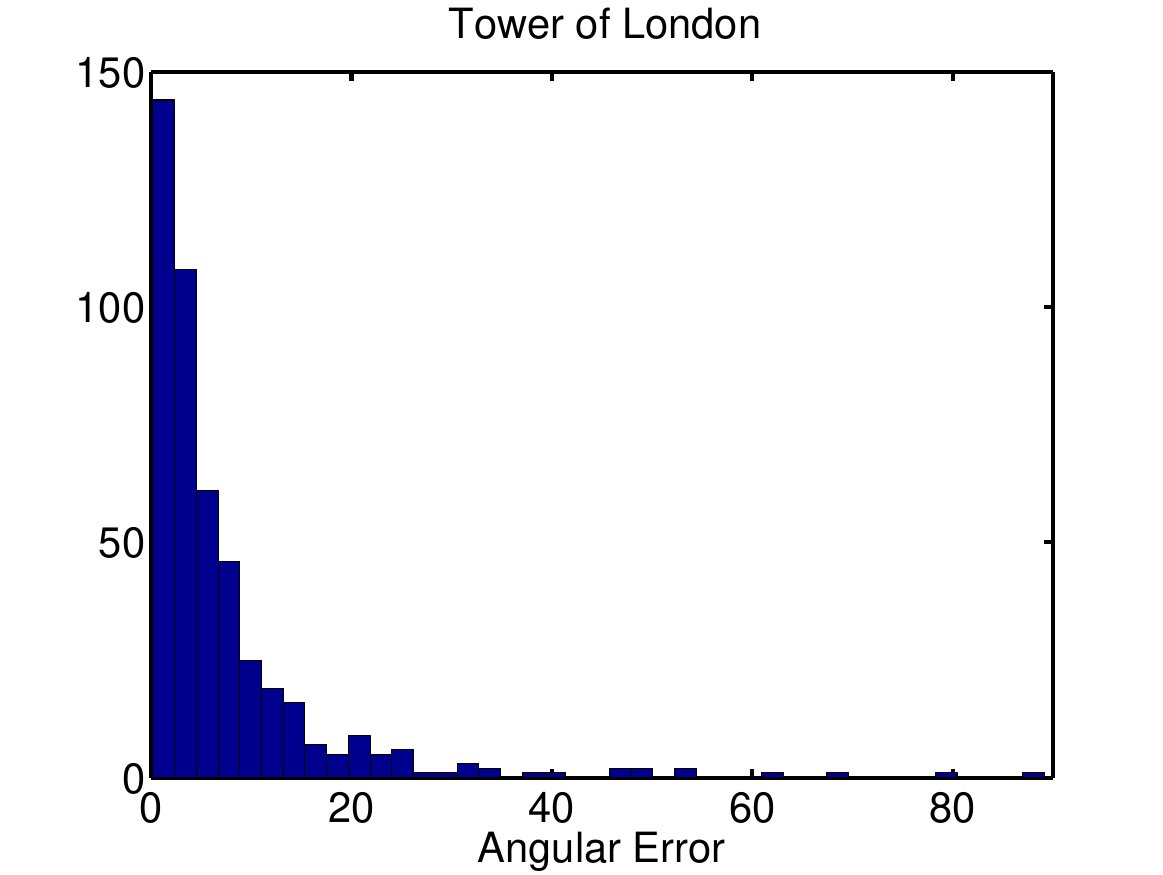}
\includegraphics[width=0.3\linewidth]{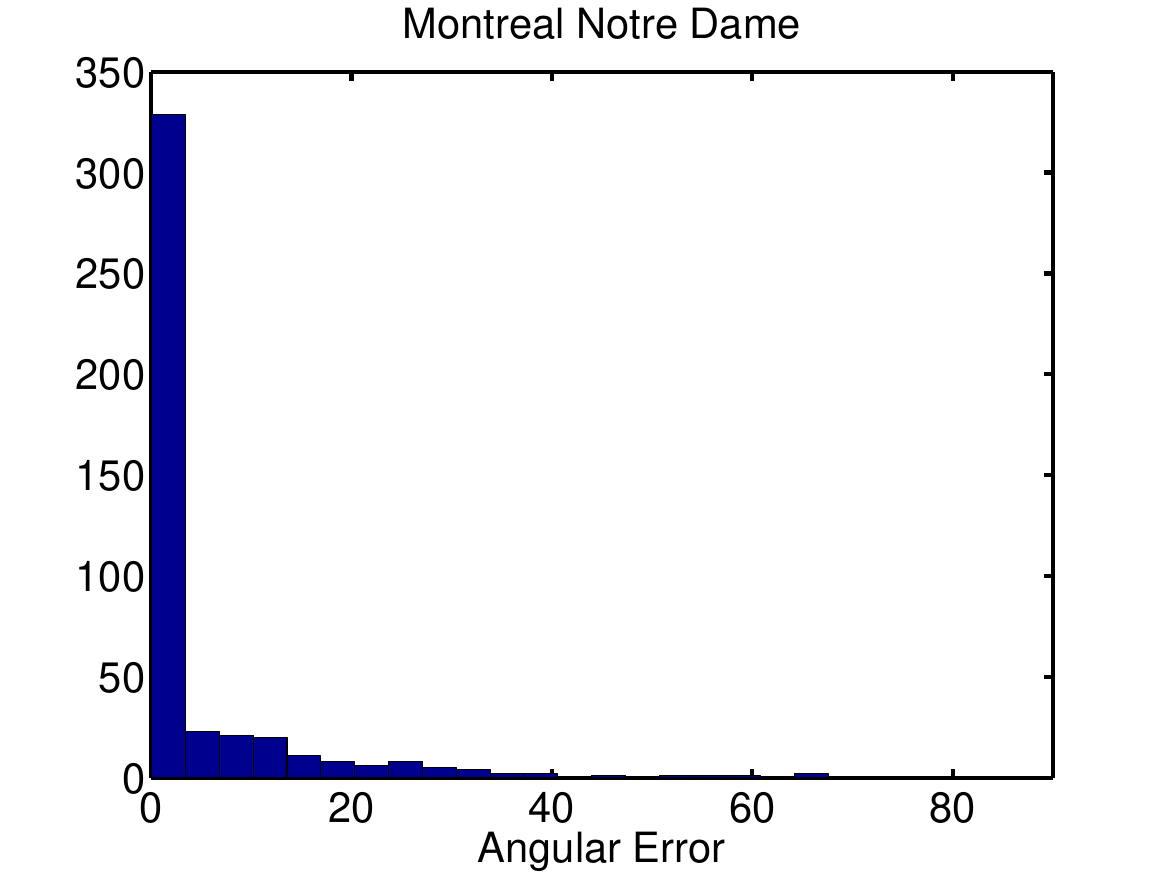}
\includegraphics[width=0.3\linewidth]{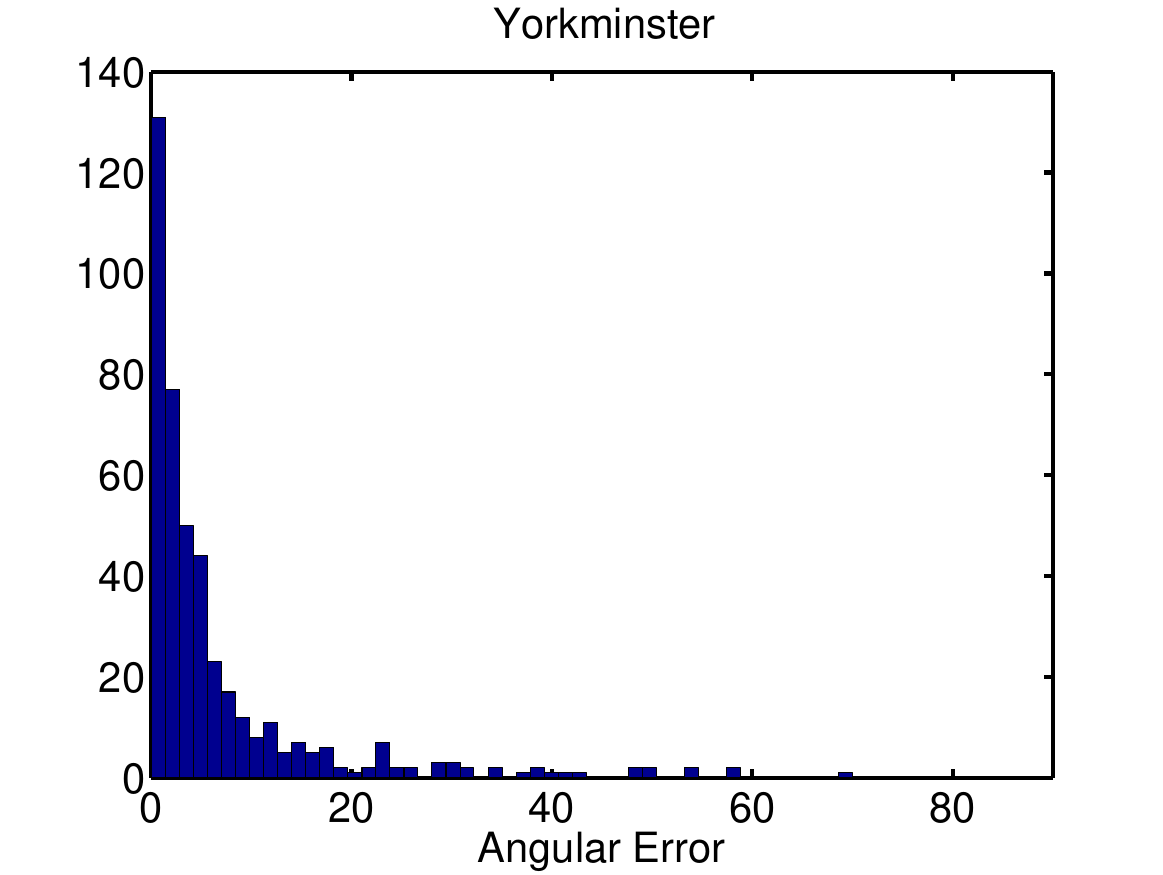} \\
\includegraphics[width=0.3\linewidth]{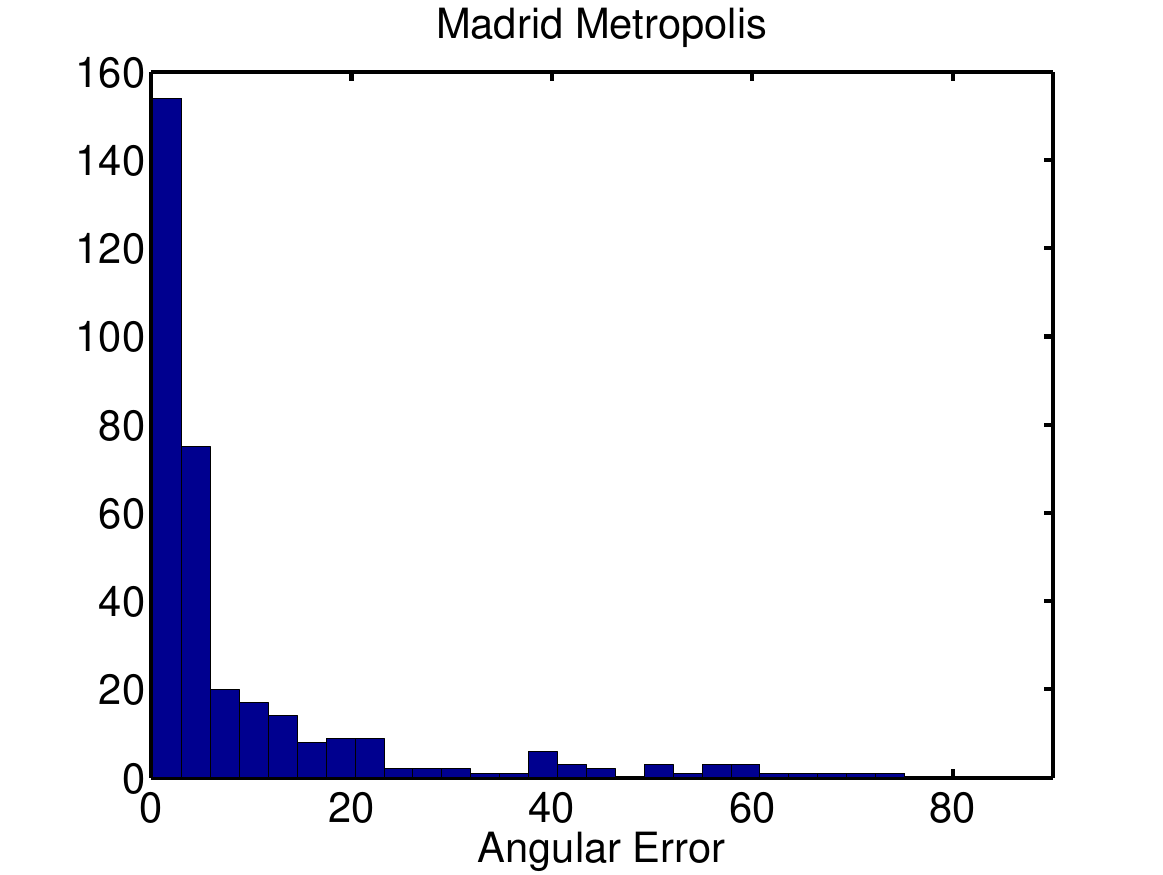}
\includegraphics[width=0.3\linewidth]{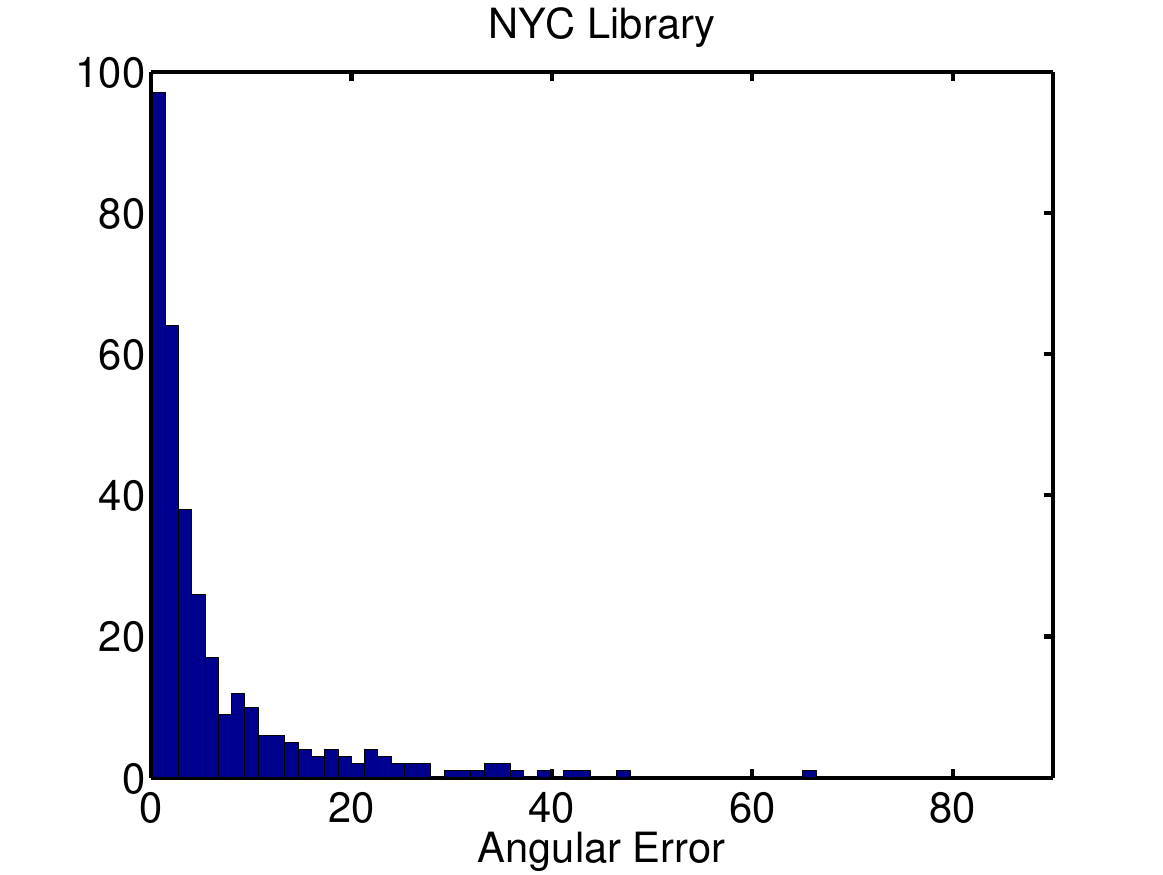}
\includegraphics[width=0.3\linewidth]{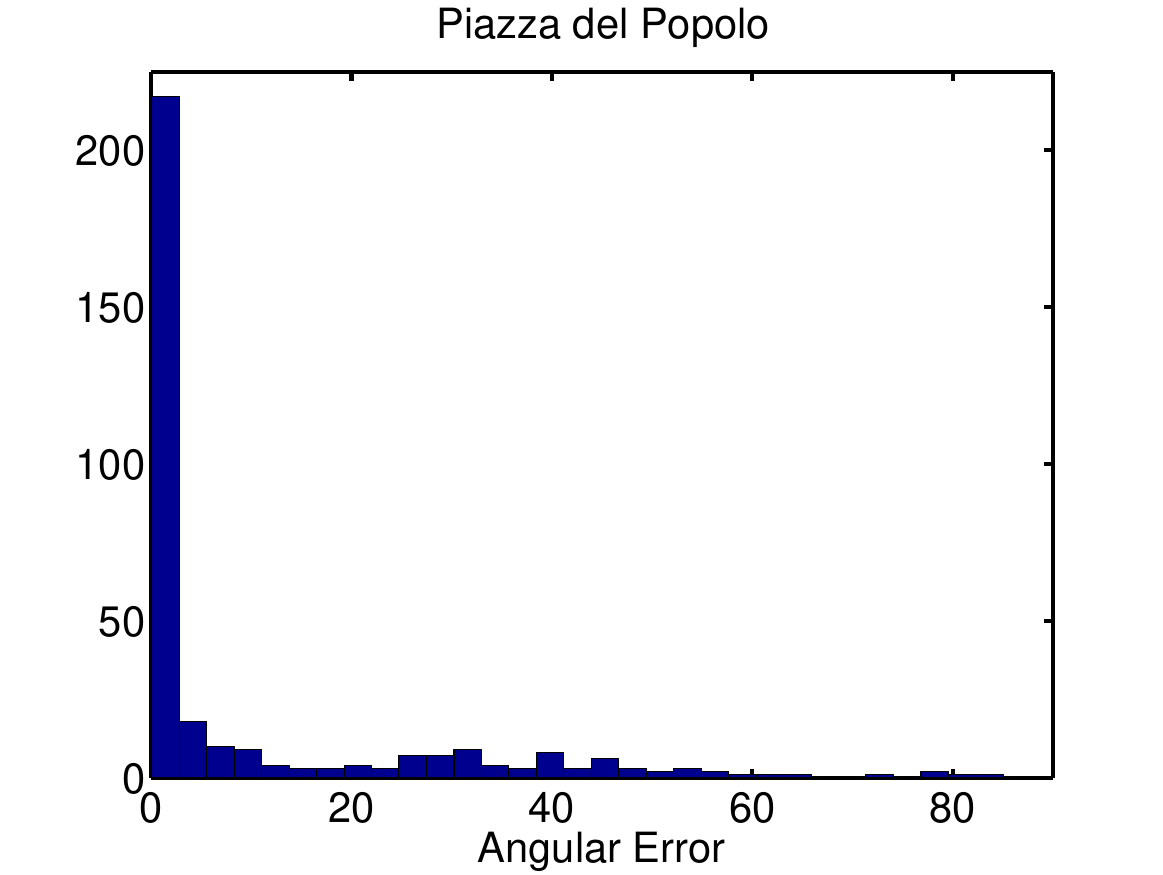} \\
\caption{Histograms of the angular errors obtained by \textsc{R-GoDec} on the largest datasets of Table 1.}
\label{histogram}
\end{figure*}

Results are shown in Table \ref{snav_table}, which reports the median angular error and the execution cost of several algorithms. The Castle-P* sequences are taken from \cite{real_data}, while the remaining datasets are taken from \cite{Snavely14}. The histograms of the errors obtained by \textsc{R-GoDec} are reported in Figure \ref{histogram}.

Neither EIG, SDP and \textsc{OptSpace} nor Weiszfeld and LUD are applicable in practical scenarios, since they do not satisfy the requirements of an efficient robust scheme.
The first three achieve the highest errors since they are not robust to outliers, whereas the last two show resilience to outliers (to variable degrees) but they have the highest execution time.

The remaining algorithms solve the rotation synchronization problem while ensuring robustness and efficiency at the same time, to different extents. In particular, L1-IRLS achieves the highest accuracy, while  \textsc{R-GoDec} and \textsc{Grasta} achieve an accuracy lower than L1-IRLS, albeit comparable, in substantially less time.
As for EIG-IRLS, results are comparable to L1-IRLS in most datasets both in terms of accuracy and execution time.
\textsc{R-GoDec} turns out to be the fastest solution among all the robust ones. 

\begin{table*}
\centering
\ra{1.25}
\caption{Median angular errors [degrees] and execution times [seconds] for several algorithms on some standard image sequences. The number of images (n) and the percentage of missing relative rotations (miss) are also reported for each dataset.}
\label{snav_table}
\medskip
\resizebox{\linewidth}{!}{
\begin{tabular}{lrrrrrrrrrrrrrrrrrrrrrrrrrrrrrr}
\toprule
 &&
\multicolumn{2}{c}{}  &&   
\multicolumn{2}{c}{EIG} &&   
\multicolumn{2}{c}{\textsc{OptSpace}} &&   
\multicolumn{2}{c}{SDP} && 
 \multicolumn{2}{c}{Weiszfeld} &&   
 \multicolumn{2}{c}{LUD}  &&   
 \multicolumn{2}{c}{EIG-IRLS} && 
 \multicolumn{2}{c}{L1-IRLS} &&  
 \multicolumn{2}{c}{\textsc{R-GoDec}} &&
\multicolumn{2}{c}{\textsc{Grasta}}  \\
%\cmidrule{3-4} 
\cmidrule{6-7} \cmidrule{9-10} \cmidrule{12-13} \cmidrule{15-16}
\cmidrule{18-19} \cmidrule{21-22} \cmidrule{24-25} \cmidrule{27-28} \cmidrule{30-31}
&& n & miss && err & t && err & t && err & t && err & t 
&& err & t && err & t && err & t && err & t && err & t \\
\midrule
Vienna Cathedral      && 918 & 75 && 5.96 & 4.9  && 5.34 & 20   && 6.15 & 5090 && 3.68 & 590 && - & -        && 1.60 & 120   && 1.37 & 136  && 2.77 & 12   && 1.93 & 27   \\		  
Alamo                 && 627 & 50 && 3.16 & 2.5  && 2.92 & 10   && 3.21 & 1425 && 2.11 & 587 && - & -        && 1.18 & 113   && 1.09 & 79   && 1.47 & 6.5  && 1.30 & 20   \\  	 
Notre Dame	          && 553 & 32 && 3.44 & 2.2  && 3.03 & 10   && 3.66 & 836  && 1.88 & 527 && - & -        && 0.74 & 93    && 0.65 & 68   && 1.03 & 4.8  && 0.75 & 9.1  \\		  
Tower of London       && 508 & 81 && 3.87 & 0.8  && 3.76 & 3.4  && 3.98 & 757  && 3.32 & 142 && - & -        && 2.78 & 24    && 2.63 & 6.3  && 4.05 & 5.9  && 3.10 & 7.6  \\		  
Montreal N.~Dame      && 474 & 53 && 2.24 & 1.1  && 1.87 & 5.2  && 2.29 & 601  && 1.15 & 301 && - & -        && 0.59 & 40    && 0.58 & 24   && 0.84 & 5.5  && 0.67 & 7.5  \\		  
Yorkminster           && 458 & 74 && 5.85 & 0.64 && 4.96 & 3.2  && 5.68 & 558  && 3.75 & 145 && 1.87 & 2416  && 1.82 & 26    && 1.69 & 6.9  && 3.13 & 9.8  && 2.06 & 8.3  \\		  
Madrid Metropolis     && 394 & 69 && 7.48 & 0.57 && 6.70 & 2.2  && 7.42 & 388  && 5.54 & 124 && 4.56 & 1792  && 4.43 & 17    && 1.01 & 16   && 3.26 & 5.1  && 2.62 & 6.8  \\		  
NYC Library           && 376 & 71 && 5.51 & 0.39 && 5.33 & 2.1  && 5.58 & 304  && 3.68 & 102 && 1.95 & 1791  && 1.99 & 12    && 1.33 & 6.6  && 2.98 & 2.1  && 1.90 & 6.2  \\		  
Piazza del Popolo     && 354 & 60 && 3.34 & 0.53 && 3.11 & 2.2  && 3.48 & 246  && 2.27 & 139 && 1.22 & 1554  && 1.03 & 41    && 0.98 & 12   && 1.42 & 2.9  && 1.20 & 3.5  \\		
Ellis Island	      && 247 & 33 && 2.81 & 0.22 && 2.63 & 1.3  && 2.89 & 78   && 1.50 & 97  && 0.91 & 595   && 0.81 & 29    && 0.57 & 6.8  && 1.01 & 1.7  && 0.77 & 2.8  \\		  
Castle-P30            && 30  & 55 && 1.18 & 0.07 && 1.19 & 0.11 && 1.26 & 0.01 && 0.53 & 2.6 && 0.46 & 14.9  && 0.28 & 0.49  && 0.59 & 0.10 && 0.67 & 0.06 && 0.28 & 0.30 \\	  
Castle-P19            && 19  & 58 && 1.25 & 0.07 && 1.32 & 0.08 && 1.37 & 0.01 && 0.56 & 1.1 && 0.81 & 1.8   && 0.33 & 0.48  && 0.91 & 0.08 && 0.54 & 0.05 && 0.50 & 0.19 \\        
% Entry-P10      && & && & &&  & && & && &  && & && & &&  & && & && &   \\ 	
% Fountain-P11   && & && & &&  & && & && &  && & && & &&  & && & && &   \\
% Herz-Jesu-P25  && & && & &&  & && & && &  && & && & &&  & && & && &   \\
% Herz-Jesu-P8   && & && & &&  & && & && &  && & && & &&  & && & && &   \\
\bottomrule
\end{tabular}
}
\end{table*}

\section{Conclusion}
\label{conclusion}

After reviewing the literature on low-rank and sparse matrix decomposition and conceiving
a new algorithm (\textsc{R-GoDec}) that addresses RPCA and MC together, 
we showed how the \ARE problem, in the presence of missing data and ouliers, 
can be formulated as a low-rank and sparse decomposition.
Then we modified \textsc{R-GoDec} to exploit the block structure of our measures
and we applied it to the solution of the \ARE problem. Experiments on simulated and 
real data show that \textsc{R-GoDec} is the fastest among the robust methods, 
while demonstrating a sufficient resilience to outliers.
Our novel formulation opens the way to the application of matrix decomposition techniques to structure-from-motion and multiple point-set registration, since -- in principle -- any algorithm able to perform RPCA and MC simultaneously can be used in place of \textsc{R-GoDec}.

\bibliographystyle{elsarticle-num}
\bibliography{references}

\end{document}